\documentclass{article}

\usepackage[main, final]{neurips_2025}

\usepackage[utf8]{inputenc} 
\usepackage[T1]{fontenc}    
\usepackage{hyperref}       
\usepackage{url}            
\usepackage{booktabs}       
\usepackage{amsfonts}       
\usepackage{nicefrac}       
\usepackage{microtype}      
\usepackage[dvipsnames]{xcolor}         
\usepackage{colortbl}
\usepackage{graphicx}
\usepackage[font=small,labelfont=bf]{caption}
\usepackage{wrapfig}
\usepackage{amssymb}
\usepackage{makecell}
\usepackage{tabularx}
\usepackage{enumitem}
\usepackage{algorithm}
\usepackage{algpseudocode}
\usepackage{mathrsfs}

\usepackage{amsmath}
\newcommand*{\vocab}{\mathbf{\Omega}}
\newcommand*{\vocabsize}{\Omega}
\usepackage{amsthm}
\newtheorem{theorem}{Theorem}

\newtheorem{definition}[theorem]{Definition}

\DeclareMathOperator*{\E}{\mathbb{E}}

\newcommand*{\green}[1]{\textcolor{ForestGreen}{#1}}
\newcommand*{\red}[1]{\textcolor{BrickRed}{#1}}

\usepackage[most]{tcolorbox}
\newtcolorbox{promptbox}[1][]{%
  enhanced, breakable,
  colback=gray!2, colframe=gray!55,
  boxrule=0.4pt, sharp corners,
  left=6pt, right=6pt, top=6pt, bottom=6pt,
  title=\bfseries #1
}

\usepackage{comicneue}

\title{Correlation Dimension of\\Auto-Regressive Large Language Models}

\author{%
Xin Du\\
Waseda University\\
\texttt{duxin@aoni.waseda.jp} \\
\And Kumiko Tanaka-Ishii \\
Waseda University\\
\texttt{kumiko@waseda.jp} \\
}

\begin{document}

\maketitle

\begin{abstract}

Large language models (LLMs) have achieved remarkable progress in natural
language generation, yet they continue to display puzzling behaviors—such as
repetition and incoherence—even when exhibiting low perplexity. This
highlights a key limitation of conventional evaluation metrics, which
emphasize local prediction accuracy while overlooking long-range structural
complexity.  We introduce correlation dimension, a fractal-geometric measure
of self-similarity, to quantify the epistemological complexity of text as
perceived by a language model. This measure captures the hierarchical
recurrence structure of language, bridging local and global properties in a
unified framework.  Through extensive experiments, we show that correlation
dimension (1) reveals three distinct phases during pretraining, (2) reflects
context-dependent complexity, (3) indicates a model's tendency toward
hallucination, and (4) reliably detects multiple forms of degeneration in
generated text.  The method is computationally efficient, robust to model
quantization (down to 4-bit precision), broadly applicable across
autoregressive architectures (e.g., Transformer and Mamba), and provides
fresh insight into the generative dynamics of LLMs.
    
\end{abstract}

\section{Introduction}

Latest advances in large language models (LLMs) have demonstrated sophisticated
capabilities, including mathematical reasoning and planning. These abilities
suggest that LLMs internally process information through complex, nonlinear, and
potentially hierarchical mechanisms \citep{allen2023physics}, despite operating
through simple, token-by-token predictions. Understanding precisely how LLMs
achieve such macroscopic behaviors from microscopic predictive steps remains an
important open question, essential for grasping the full potential and
limitations of these models. Although previous research shows that minimizing
next-token prediction loss---perplexity---is theoretically powerful
\citep{malach2024auto}, it still falls short in fully explaining unexpected
model behaviors, such as hallucinations and repetitive, bland outputs, even when
perplexity values are low. Thus, new tools for characterizing the complex
behavior of LLMs are urgently needed.

Current evaluation approaches broadly fall into two categories. The first
includes methods based on local textual properties, such as lexical or syntactic
statistics (e.g., $n$-gram frequencies). These metrics are intuitive and
interpretable but often fail to capture semantic ambiguities or deeper
structural complexities. The second category consists of global metrics, such as
mean perplexity or semantic similarity measures. While these methods provide
comprehensive quantitative evaluations, they frequently lack interpretability
and connection to underlying local textual properties. This divide between local
interpretability and global comprehensiveness reflects the broader challenge of
bridging microscopic (token-level) and macroscopic (long-range, structural)
perspectives of LLM behavior.

In this work, we propose {\em correlation dimension}, a measure drawn from
fractal geometry and dynamical systems theory, to bridge this gap. Correlation
dimension quantifies self-similarity---a fundamental characteristic of complex
systems that exhibit invariant patterns across scales. Originally developed for
analyzing deterministic chaotic systems \citep{grassberger1983characterization},
correlation dimension has since been successfully adapted to stochastic
processes and real-world complex phenomena
\citep{sauer1991embedology,lacasa2013correlation}. Applied to language models,
correlation dimension effectively quantifies the intrinsic complexity and
recurrence structure of generated texts as perceived by the model. For instance,
texts with randomly shuffled words appear highly complex and yield high
dimensionality, whereas simple repetitive patterns exhibit low dimensionality.

We specifically propose computing correlation dimension using sequences of
next-token log-probability vectors, which are readily available for any
autoregressive language model. Recurrences are defined via the Euclidean
distance between these vectors. Unlike perplexity, correlation dimension
reflects deeper structural properties of the text generation process. For
example, a model experiencing mode collapse might produce contextually
irrelevant but statistically plausible text, exhibiting low perplexity yet high
correlation dimension. Conversely, models lacking adequate memory to handle
long-range dependencies might show low dimensionality despite high perplexity,
clearly indicating their limited structural comprehension.

Correlation dimension serves as both a practical and theoretically grounded
metric. It requires little computational overhead beyond perplexity calculations
and runs at inference time, making it easy to integrate into existing inference
infrastructures such as \texttt{vllm}. At a theoretical level, it provides
nuanced insights into model behavior, revealing the long-range complexity
structures that perplexity alone cannot uncover. Practically, it offers a
straightforward yet powerful method to evaluate model robustness, indicate
potential problems such as hallucination, and capture subtle forms of
degeneration such as incoherence, and blandness.

Throughout the paper, we illustrate how correlation dimension inherently
connects local recurrence structures and global textual complexity (Section
\ref{sec:language}), provides insights into intrinsic properties of texts and
model behavior (Section \ref{sec:charac-llm}), and effectively detects
degeneration issues in long-text generation (Section \ref{sec:degenerate}).
Overall, our approach naturally integrates interpretability and
comprehensiveness, offering a robust metric grounded in the fundamental
properties of complex dynamical systems.

\section{Related Works}
\label{sec:related}

\subsection{Statistical Self-Similar Phenomenon in Language}

Self-similarity is a fundamental property observed in various complex systems,
characterized by invariant patterns across multiple scales. A special class of
self-similarity, known as scale-free properties, refers specifically to patterns
that are consistent when viewed at different scales
\citep{thurner2018introduction}. Unlike precise scale-free structures defined
mathematically (e.g., the Koch snowflake), {\em statistical self-similarity}
denotes approximate scale invariance identified through statistical analysis of
empirical data \citep{tanaka2021statistical}.

Statistical self-similarities have been widely documented in natural language,
manifesting through phenomena such as Zipf's law
\citep{zipf1949human,mandelbrot1953informational,piantadosi2014zipf} for word
frequencies, Herdan-Heaps's law \citep{herdan1966advanced,heaps1978information}
for vocabulary growth, and long-range correlations
\citep{ebeling1994,ebeling1995,Altmann2012,tanaka2018taylor,de2023multifractal}
for word occurrence or word rank across text spans.

Previous approaches have explored statistical self-similarity and fractal
dimension at the abstract semantic level, extending beyond traditional
lexical-based metrics (e.g., word frequency). For instance,
\cite{doxas2007dimensionality, doxas2010dimensionality} utilized latent semantic
analysis (LSA) embeddings of text paragraphs \citep{deerwester1990indexing} and
discovered geometric self-similarity (fractal structures) within semantic
spaces, estimating fractal dimensions approximately between 8 and 20.
\cite{ribeiro2023fractal} also observed fractal patterns in word vectors with a
Box-counting dimension of about 20.
These findings are based on static word
vectors that do not vary with context, and therefore have limited utility for
characterizing language models. 

Recently, \cite{du2024corrdim} investigated the self-similarity of texts using
autoregressive LLMs, and found consistent semantic self-similarity across
multiple languages.  However, the linguistic significance of the self-similarity
and the explicit role of language models in capturing such phenomena remains
unexplored.  In contrast, this paper proposes a new method tailored for
analyzing LLMs, and provides a range of insights into the generative dynamics of
LLMs.

\paragraph{Characterizing LLMs via Self-Similarity}

Previous studies have evaluated language models via statistical self-similarity.
\cite{takahashi2019evaluating} compared generated texts from language models to
natural texts across five scale-free properties, revealing significant
deviations in scaling exponents. Building on this, \cite{meister2021language}
introduced quantitative statistical tests to measure such deviations
systematically and examined how different sampling strategies (e.g., nucleus
sampling vs. beam search) affect self-similarity.  \cite{chen2025l2m} set up a
mutual-information scaling law for long-context language modeling.

Notably, \cite{alabdulmohsin2024fractal} recently measured the fractal dimension
and Hurst exponent of cumulative log-perplexity series, demonstrating that these
scaling exponents correlate with downstream performance of LLMs. However, their
approach primarily captures long-range dependencies in the model's predictive
error sequences. In contrast, our method (detailed in Section
\ref{sec:language}) examines the intrinsic generative recurrence structure
inherent in language itself, uncovering fundamentally different insights about
the model behavior and text complexity.

\section{Recurrence Structure and Correlation Dimension of Language}
\label{sec:language}

\begin{figure}
    \centering
    \small
    \begin{minipage}[b]{0.44\textwidth}
        \centering \small
        \includegraphics[width=\textwidth]{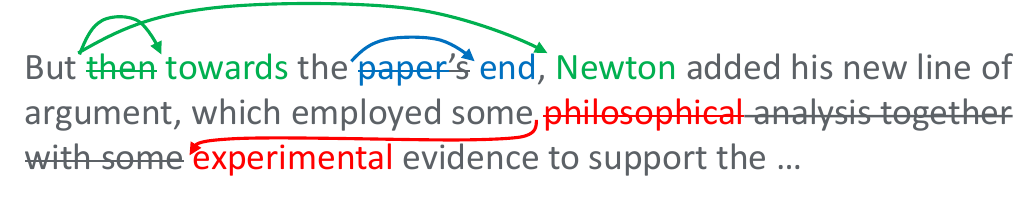} \\
        (a) Local skips
        \vskip 1em
        \includegraphics[width=\textwidth]{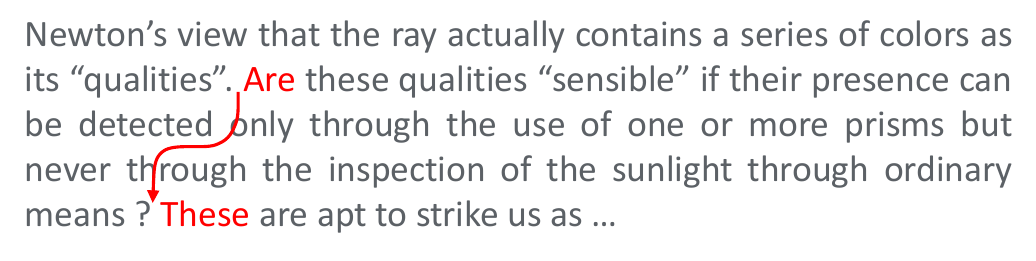} \\
        (b) Long-range skips (whole sentences)
    \end{minipage}
    \begin{minipage}[b]{0.24\textwidth}
        \centering \small
        \includegraphics[width=\textwidth]{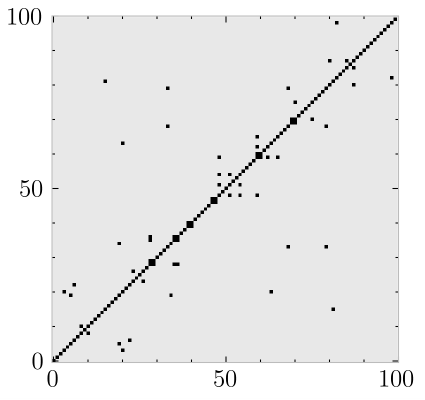} \\
        \vskip 1.5em
        (c) Recurrence plot
    \end{minipage}
    \begin{minipage}[b]{0.29\textwidth}
        \centering \small
        \includegraphics[width=\textwidth]{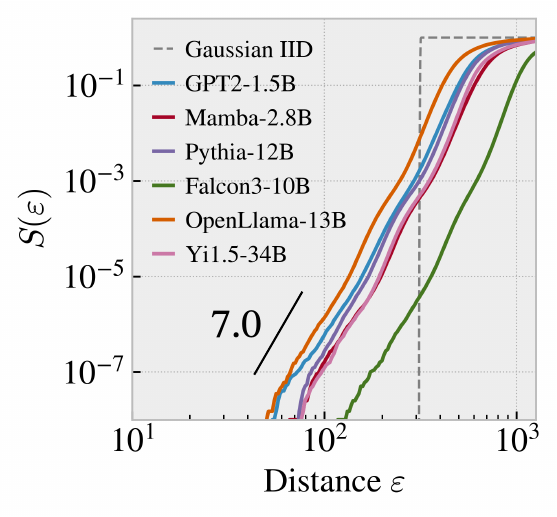} \\
        \vskip -0.25em
        (d) Correlation integral curves
    \end{minipage}
    \caption{
        Experimental results on the ``Newton's Philosophy'' article of the Stanford
        Encyclopedia of Philosophy.
        (a-b) Examples of textual skips as predicted by the Pythia-12B model, illustrating both local and long-range recurrences in the text.
        (c) Segment of the recurrence plot for the log-probability vectors generated by the model; black dots indicate pairs of points within a specified distance threshold.
        (d) Correlation integral curves for six pre-trained language models (solid lines), compared to i.i.d. Gaussian noise in $\mathbb{R}^{50000}$ (dashed lines).
    }
    \label{fig:newton}
\end{figure}

Natural language exhibits self-similarity across various linguistic scales,
ranging from morphological, lexical, and syntactic to semantic structures and
long-range dependencies. Unlike mathematically precise fractals, real-world
self-similarity is typically approximate and statistical, manifesting through
recurrent statistical patterns rather than exact repetitions.

The {\em correlation dimension} provides a quantitative framework to
measure such statistical self-similarity in sequences by analyzing their {\em recurrence}
structure. A recurrence is defined as an event where the trajectory of a
system approximately revisits a previous state within a predefined distance
threshold $\varepsilon$. As $\varepsilon$ increases, more recurrences
naturally emerge. Self-similar systems typically follow a power-law
relationship between the recurrence frequency $S(\varepsilon)$ and the
distance threshold $\varepsilon$, formally defined as follows:

\begin{definition}[Correlation dimension \citep{grassberger1983characterization}]
    \label{def:corrdim}
    Given an infinite sequence $\{x_t\}_{t=1}^\infty$ in a metric space (e.g., $\mathbb{R}^D$),
    its correlation dimension $d$ is the exponent characterizing the scaling behavior
    of the correlation integral
    $S(\varepsilon)$:
    \begin{equation}
        S(\varepsilon)\quad \propto \quad \varepsilon^d \quad \text{as} \quad \varepsilon\to 0,
        \label{eq:corrint}
    \end{equation}
    where the correlation integral $S(\varepsilon)$ is defined as
    the frequency of point pairs separated by distances less than $\varepsilon$:
    \begin{equation}
        S(\varepsilon) = \lim_{t\to\infty} \frac{2}{t(t-1)}
        \sum_{1\leq i<j\leq t} 1\{\lVert x_i - x_j \rVert < \varepsilon\}.
    \end{equation}
    Here, $\lVert \cdot\rVert$ denotes the Euclidean norm, and $1\{\cdot\}$ is the
    indicator function that equals 1 if the condition is true and 0 otherwise.
\end{definition}

Correlation dimension was originally developed to characterize deterministic
attractors (e.g., H\'enon maps \citep{henon2004two}) but subsequently
generalized for analyzing stochastic processes such as fractional Brownian
motion \citep{flandrin1989spectrum} and complex networks
\citep{lacasa2013correlation}.

\paragraph{Correlation Dimension Applied to Language.} Representing text as a
sequence of numerical vectors allow the identification and measurement of
recurrences and subsequently their correlation dimension. However, selecting a
meaningful, stable, and interpretable representation for natural language is
nontrivial.

We propose utilizing the Euclidean distance between {\em logarithmic} next-token probability
vectors derived from language models. Specifically, the log-probability vector at time $t$,
denoted $x_t$, is calculated as:
\begin{equation}
    x_t(\omega) = \log \mathrm{P}_\theta(\omega_t=\omega | \omega_{t-c},\cdots,\omega_{t-1})
    \quad \forall \omega\in\vocab,
    \label{eq:xt}
\end{equation}
where $\omega_t$ represents the token at position $t$, $\mathrm{P}_\theta$
represents the model-predicted probability ($\theta$ specifies the model), and
$\vocab$ is the vocabulary set.  Unless otherwise specified, the context length
$c$ is set to infinity, i.e., no context length limit; its impact is further
examined in Section \ref{sec:ctxlen} and discussed in Appendix \ref{sec:estimation}.

\paragraph{Textual Skips as Recurrences.} Recurrences in next-token
log-probability vectors can be interpreted as {\em textual skips}. If two states
$x_t$ and $x_s$ ($s<t$) are close, the text segment $[s, t)$ could theoretically
be omitted without significantly altering subsequent text generation. Skips
occur at multiple scales, as illustrated in Figure \ref{fig:newton}(a-b): from
local (single-word skips) to global (sentence-level skips). Smaller distance
thresholds ($\varepsilon$) identify local skips, while larger thresholds detect
longer-range skips. The hierarchy aligns naturally with Chomsky's generative
grammar \citep{chomsky2014aspects}, where such skips may correspond to omitted
subtrees within the hierarchical structure of language.

Figure \ref{fig:newton}(c) provides an example recurrence plot based on
log-probability vectors from the text {\em Newton's Philosophy}
\citep{sep-newton-philosophy}, clearly visualizing recurrences at a certain
threshold. The correlation integrals $S(\varepsilon)$, illustrated in Figure
\ref{fig:newton}(d), demonstrate near-linear scaling for multiple pre-trained
language models (GPT2 \citep{radford2019language}, Pythia
\citep{biderman2023pythia}, Falcon3 \citep{Falcon3}, OpenLLaMA
\citep{openlm2023openllama}, Yi1.5 \citep{yi}, and Mamba \citep{gu2024mamba}),
yielding a consistent correlation dimension around 7. In contrast, Gaussian
random vectors in equivalent dimensional spaces exhibit distinct different
scaling behavior, underscoring language-specific recurrence structures.

\subsection{Sufficiency of Next-Token Log-probabilities}

The next-token log-probability vectors represent only partial information of a
language model's full state, defined theoretically as the distribution over all
future token sequences. Thus, one might question the sufficiency of next-token
probabilities alone for characterizing long-range text generation.

Time-delayed embeddings, motivated by Takens' embedding theorem
\citep{takens1980detecting} and its stochastic extensions
\citep{sauer1991embedology,stark2003delay}, are common methods for
reconstructing full states from partial observations. Such embeddings
concatenate multiple sequential log-probability vectors, $y_t=[x_t; x_{t+1};
    \dots; x_{t+k}]$, potentially capturing higher-order dependencies.

However, empirical observations reveal negligible differences in the correlation
dimension when comparing simple next-token log-probability vectors ($k=1$) to
embeddings with higher orders ($k>1$), after accounting for the noise inevitably
introduced in the delayed embeddings. This somewhat surprising result suggests
that single-step log-probability vectors inherently encode significant long-term
structural information about language evolution. This phenomenon mirrors
findings in knowledge distillation \citep{hinton15distill}, where single-step
probability distributions effectively summarize model knowledge. Additional
details and analysis are provided in Appendix \ref{sec:delay}.

\subsection{Empirical Convergence of Correlation Dimension}

\begin{figure}[t]
    \centering
    \begin{minipage}{0.8\linewidth}
        \centering \small
        \includegraphics[width=\linewidth]{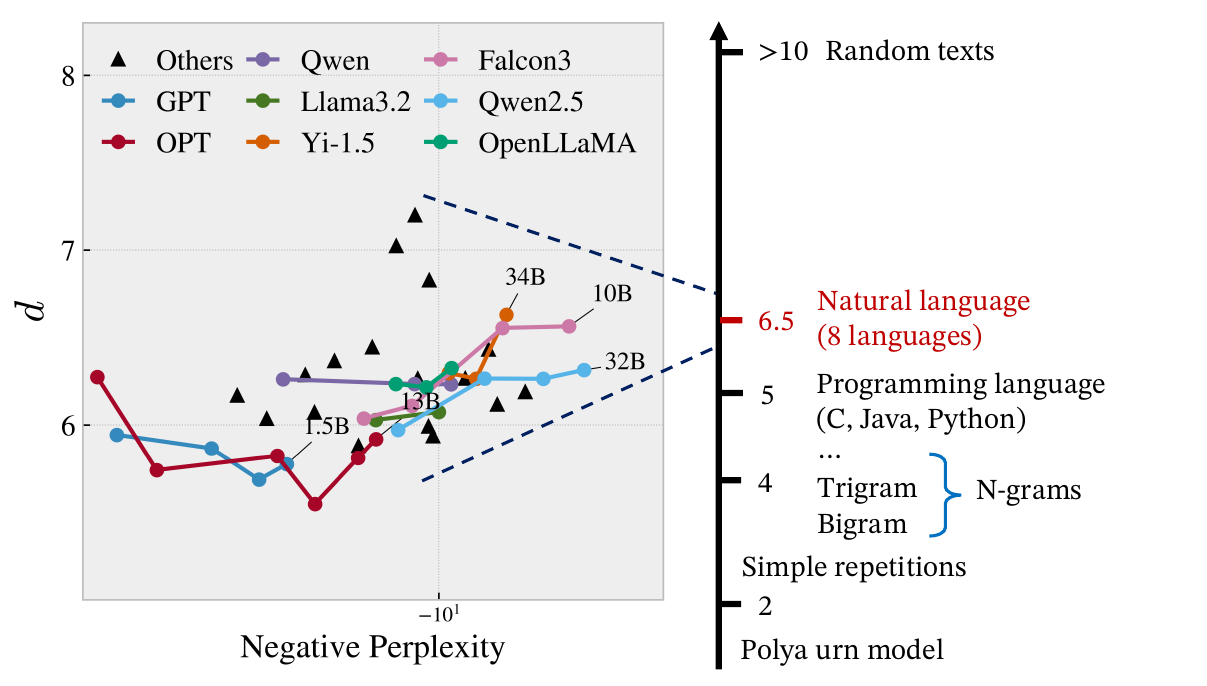}
    \end{minipage}
    \caption{Left: Mean correlation dimension of the SEP dataset measured with
        various pre-trained LLMs.
        Right: A conjectured spectrum of correlation dimension for different
        types of texts / language models.}
    \label{fig:modelsize}
\end{figure}

Considering an ideal ``perfect'' language model capable of precisely predicting
next-token distributions, its correlation dimension would reflect intrinsic
textual complexity. However, slight imperfections in predictions significantly
affect correlation dimension measurements, particularly due to rare outcomes
influencing the measure disproportionately compared to typical loss calculations.
Precisely, let
$l(\theta)=\E_{\omega_{\leq t}\sim \mathbb{P}} \bigl[\log \mathrm{P}_\theta
        (\omega_{t}|\omega_{\leq t-1})\bigr]$ be the cross-entropy loss function of a
language model parameterized by $\theta$, where $\mathbb{P}$ and
$\mathrm{P}_\theta$ represent the empirical distribution and the model-predicted
distribution, respectively, and $\omega_{\leq t}=[\omega_1,\cdots,\omega_{t}]$
represents the sequence of tokens until time $t$. Then, we can see that the
contribution of rare outcomes $\omega$ to the total loss is very small,
proportional to its frequency:
\begin{equation}
    \dfrac{\partial\, l(\theta)}{\partial\, \log \mathrm{P}_\theta
        (\omega|\omega_{\leq t-1})} = \mathbb{P} (\omega|\omega_{\leq t-1}) \qquad
    \forall\omega\in\vocab.
    \label{eq:loss}
\end{equation}
In other words, the log-probability vectors at rare words $\omega$ may
vary significantly, without affecting the loss function, but this variation
will be reflected in the correlation dimension regardless of the rareness
of the word.

Nonetheless, empirical experiments indicate convergence in correlation
dimensions of well-trained language models to a narrow range as perplexity
decreases. Figure \ref{fig:modelsize} illustrates correlation dimension for
various pre-trained LLMs across the Stanford Encyclopedia of Philosophy (SEP)
\citep{sep} dataset which is summarized in Appendix \ref{app:sep}. As the
perplexity decreases, correlation dimensions stabilize around a consistent value
near 6.5.  This empirical convergence underscores correlation dimension's
reliability as a stable, interpretable metric for evaluating the structural
complexity of natural language and the performance of language models.

The right half of Figure \ref{fig:modelsize} presents the spectrum of
correlation dimension values across various types of texts and statistical
processes. Randomly shuffled texts exhibit high correlation dimensions,
typically above ten, whereas self-reinforcing processes such as the Polya urn
model \citep{mahmoud2008polya} display much lower values, below two. $n$-gram
processes with small $n$ also yield lower correlation dimensions compared to
natural language. Moreover, programming languages (Python, Java, and C) show a
consistent correlation dimension around 5. We provide the full results in
Appendix \ref{sec:other-natural-languages} (other natural languages) and
\ref{sec:programming-languages} (programming languages).

\section{Characterizing Language Models Using Correlation Dimension}
\label{sec:charac-llm}

Estimating the correlation dimension of a text using language models raises
critical questions regarding the reliability and interpretability of this
measure when the underlying model is imperfect or insufficiently trained. Ideally,
with a perfect language model, the correlation dimension would represent the
intrinsic complexity of the text itself. However, real-world language models
inevitably contain imperfections, prompting us to investigate whether the
correlation dimension remains meaningful under such conditions.

In this section, we demonstrate that the correlation dimension is a robust
measure of the {\em perceived complexity} of texts by LLMs, even when the LLMs
are imperfect or insufficiently trained.  Specifically, we explore three key
aspects: (1) how the correlation dimension reflects the intrinsic complexity of
a text as determined by its underlying hierarchical structure (Section
\ref{sec:lin-tegmark}); (2) how the correlation dimension is influenced by
constraints on the contextual information available to the model (Section
\ref{sec:ctxlen}), and (3) how three distinct stages emerge in the training
process of LLMs as indicated by the correlation dimension, a phenomenon not
observed in perplexity (Section \ref{sec:training}).  After that, we present a
case study of occasions where significant divergence of correlation dimension is
observed between different models on knowledge-intensive texts, and how this
divergence indicates LLMs' hallucination behavior (Section
\ref{sec:hallucination}).

\subsection{Correlation Dimension and Textual Complexity}
\label{sec:lin-tegmark}

\begin{figure}[tbp]
    \centering
    \small
    \begin{minipage}[b]{0.45\textwidth}
        \centering\small
        \includegraphics[width=\linewidth]{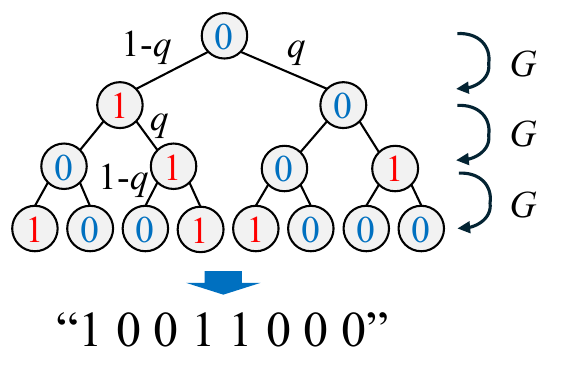}
        (a)
    \end{minipage}
    \hspace{0.05\linewidth}
    \begin{minipage}[b]{0.45\textwidth}
        \centering\small
        \includegraphics[width=\linewidth]{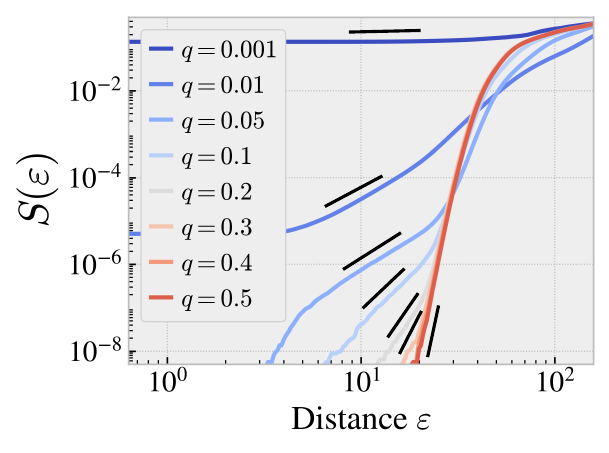}
        (b)
    \end{minipage}
    \caption{
        (a) Illustration of the Lin-Tegmark grammar generation process governed by the
        transition matrix $G$, parameterized by $q$. (b) Correlation
        integral curves for texts generated with different $q$ values, measured using
        the OpenLLaMA-13B model.
    }
    \label{fig:lin-tegmark}
    \vskip -1em
\end{figure}

To analyze how correlation dimension captures inherent textual complexity, we
generate sequences using the Lin-Tegmark grammar \citep{lin2017tegmark}, a
probabilistic context-free grammar defined over a binary alphabet $\{0, 1\}$.
The grammar's complexity is parameterized by a single Bernoulli parameter $q\in
[0,1]$, as shown in Figure~\ref{fig:lin-tegmark}(a), controlling the mutual
information decay between tokens:
$G = \begin{bmatrix}
    q   & 1-q \\
    1-q & q
\end{bmatrix}$,
where $G_{ij}$ denotes the probability of generating a child node valued $j$
given a parent node valued $i$. The resultant text complexity varies from highly
predictable sequences (as $q$ approaches 0 or 1) to highly unpredictable
sequences (as $q$ approaches 0.5).

Figure~\ref{fig:lin-tegmark}(b) illustrates the correlation integral curves
computed by the OpenLLaMA-13B model for different values of $q$. As $q$
increases towards 0.5, the correlation dimension increases markedly from near
zero to values above ten, demonstrating its sensitivity to textual complexity.
Despite the nonlinear curves due to the model not being explicitly trained on
this grammar, the correlation dimension consistently captures the inherent
complexity defined by the grammar.

\subsection{Effect of Contextual Constraints on Correlation Dimension}
\label{sec:ctxlen}

We next explore how limiting contextual access affects the correlation dimension
measured by pretrained language models. Varying the context length parameter $c$
in Eq.~\eqref{eq:xt} imposes restrictions on the model's available context,
directly influencing its complexity perception. A context length of $c=1$
reduces the model effectively to a bigram approximation, while longer contexts
progressively enable deeper linguistic comprehension.

\begin{figure}[tbp]
    \centering
    \small
    \begin{minipage}{0.45\linewidth}
        \centering \small
        \includegraphics[width=\linewidth]{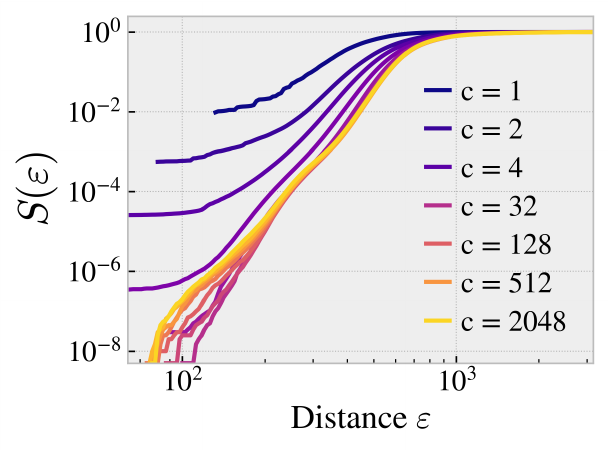}
        (a)
    \end{minipage}
    \hspace{0.05\linewidth}
    \begin{minipage}{0.45\linewidth}
        \centering \small
        \includegraphics[width=\linewidth]{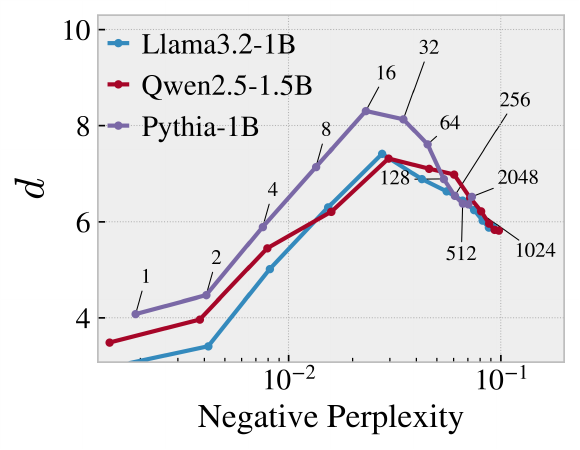}
        (b)
    \end{minipage}
    \caption{(a) Correlation integral curves on
        an article from the SEP dataset, measured by Pythia-1B model at different
        context lengths. (b) Mean correlation dimension on the SEP dataset (vertical axis),
        measured using three models at different context lengths,
        with respect to the negative perplexity (horizontal axis).
    }
    \label{fig:ctxlen}
    \vskip -1em
\end{figure}

Figure \ref{fig:ctxlen}(a) shows correlation integral curves for the Pythia-1B
model measured on the SEP dataset at varying context lengths. Figure
\ref{fig:ctxlen}(b) depicts average correlation dimensions against perplexity
for multiple models (Pythia-1B \citep{biderman2023pythia}, Qwen2.5-1.5B
\citep{qwen2.5}, and Llama3.2-1B \citep{llama3}). Notably, we observe a
two-stage pattern: initial correlation increases sharply from approximately 3 to
about 8 as context length extends to 32 tokens, followed by a gradual reduction
to around 6.5 at longer contexts.

This pattern indicates that initial increases in context length enhance the
model's perception of complexity, revealing more contextual variation.
Subsequently, the model identifies redundant contextual variations, compressing
perceived complexity and converging to a stable dimension around 6.5. This trend
consistently appears across different models, validating the correlation
dimension as a reliable metric of the complexity perceived by language models.

\begin{figure}[bp]
    \vskip -1em
    \centering
    \small
    \begin{minipage}[b]{0.49\linewidth}
        \centering \small
        \includegraphics[width=\linewidth]{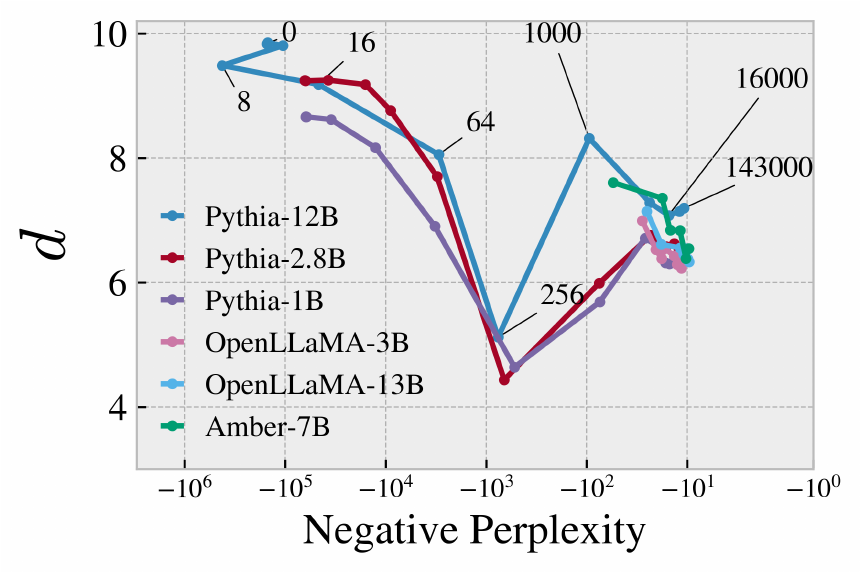}
        (a)
    \end{minipage}
    \begin{minipage}[b]{0.49\linewidth}
        \centering \small
        \includegraphics[width=\linewidth]{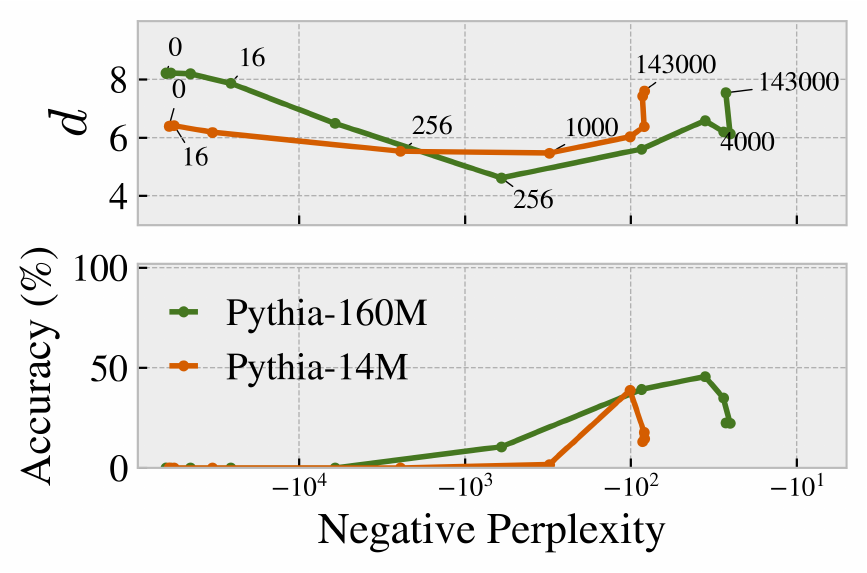}
        (b)
    \end{minipage}
    \caption{
        (a) Evolution of the mean correlation dimension for various language models on
        the SEP dataset during pre-training. (b) Notable case of two smaller models
        (Pythia-14M and 160M) that show a marked increase in correlation dimension (top)
        near the end of training, coinciding with a significant drop in in-context
        learning accuracy (bottom).
    }
    \label{fig:training}
\end{figure}

\subsection{Three-Stage Evolution in LLM Pre-Training}
\label{sec:training}

Pre-training language models typically focus on monotonically reducing perplexity,
implying a linear improvement. However, we find that correlation dimension reveals a
distinct, nonlinear evolution consisting of three stages during training.

We monitored correlation dimension across multiple checkpoints of various
models, including the Pythia family (12B, 2.8B, 1B, 160M, 14M), OpenLLaMA (3B,
13B), and Amber-7B (Figure \ref{fig:training}). The evolution distinctly unfolds
in three stages: (1) an initial rapid drop in correlation dimension due to
learning short-range (bigram-level) structures, (2) a subsequent increase as
models begin capturing longer-range dependencies, and (3) a final gradual
decline indicating improved generalization via context compression. This
three-stage evolution is clearly observed in the Pythia family, for which the
early-stage checkpoints are available; for the other models, the evolution is
verified in the last stage. Notably, these shifts occur even though perplexity
continuously decreases.

Interestingly, the third stage, indicative of improved generalization, does not
universally occur. Smaller models, like Pythia-14M and -160M, instead exhibit a
sudden increase in correlation dimension at later stages, correlating with
degraded performance in contextual learning tasks. Near the end of the training
process, the correlation dimensions of these models exhibit a sudden rise to
around 8. In the lower half, we show the accuracy of the two models in a simple
in-context learning task, where the model is asked to repeat a sequence of
symbols. A clear correlation is seen between the accuracy and the increase in
correlation dimension, suggesting the loss of generalization ability of the
model. This observation underscores the potential of correlation dimension to
indicate generalization failures and stability in language models, insights
inaccessible through perplexity alone.

\subsection{Hallucination vs. Memorization: Diverging Correlation Dimension on
    Knowledge-Intensive Texts} \label{sec:hallucination}

When processing texts containing domain-specific knowledge, LLMs may exhibit
markedly different correlation dimensions, depending on whether the model truly
recalls the knowledge or hallucinates \citep{ji2023survey}---producing
syntactically valid but factually incorrect text. A model that successfully
retrieves knowledge from memory tends to show a higher correlation dimension,
whereas a model that hallucinates typically exhibits a lower correlation
dimension on such texts.

We present a case study using the SEP article ``process-theism,'' which contains
a long list of relatively obscure scholars' names. Table
\ref{tbl:qwen-falcon-comparison} compares the correlation dimensions of the
Qwen2.5 and Falcon3 model families on the entire SEP dataset (second column) and
on the specific ``process-theism'' text (third column). Furthermore, we asked
each model to complete the name list in the text and assessed whether the model
successfully recalled the correct names or hallucinated; the results are
summarized in the fourth column (see Appendix \ref{app:hallucination} for
details).

\begin{wraptable}{r}{0.58\linewidth}
    \vskip -1em
    \small
    \caption{
        Comparison of correlation dimension between Qwen2.5 and Falcon3 on the text ``process-theism''.
    }
    \label{tbl:qwen-falcon-comparison}
    \begin{tabularx}{\linewidth}{@{\extracolsep{\fill}}lccc}
        \toprule
        \textbf{Model} & \makecell{\textbf{Normal}                              \\\textbf{text (ave.)}} & \makecell{\textbf{Knowledge-}\\\textbf{intensive text}} & \makecell{\textbf{Recalling or}\\\textbf{Hallucinating}} \\
        \midrule
        Qwen2.5-0.5B   & 5.88                      & \green{3.32} & hallucinate \\
        Qwen2.5-7B     & 6.27                      & \green{3.56} & hallucinate \\
        Qwen2.5-32B    & 6.32                      & \green{4.42} & hallucinate \\
        Falcon3-1B     & 6.03                      & \green{3.28} & hallucinate \\
        Falcon3-3B     & 6.11                      & \green{3.14} & hallucinate \\
        Falcon3-7B     & 6.55                      & \red{6.68}   & recall      \\
        Falcon3-10B    & 6.56                      & \red{8.49}   & recall      \\
        \bottomrule
    \end{tabularx}
    \vskip -1em
\end{wraptable}

As shown in the table, Falcon3-7B (6.68) and Falcon3-10B (8.49) exhibit
substantially higher correlation dimensions than the other models. Model size
alone is not the determining factor: although Qwen2.5-32B has far more
parameters, its correlation dimension remains low (4.42). A clear relationship
emerges between correlation dimension and the model's ability to recall the
correct names---every model with a correlation dimension below 5.0 produced
hallucinations in this task.

These results suggest that LLMs may internally signal whether they are
hallucinating, and that correlation dimension provides a quantitative measure of
this tendency. A possible interpretation is that recalling factual names
requires the model to engage long-range dependencies, resulting in a high
correlation dimension. In contrast, hallucination relies primarily on
format-level imitation, leading to a markedly lower correlation dimension.

\section{Correlation Dimension for Quantifying Degeneration in Text Generation}
\label{sec:degenerate}

\begin{wraptable}{r}{0.56\linewidth}
    \vskip -1em
    \small
    \caption{
        Schematic summary of various LLM evaluation methods (rows) and their ability to
        detect different types of degeneration (columns). Triangles indicate potential
        usefulness that has not been empirically validated.
    }
    \label{tbl:degeneration}
    \begin{tabularx}{\linewidth}{@{\extracolsep{\fill}}lccc}
        \toprule
        \textbf{Metric}                               & \textbf{Repetition} & \textbf{Incoherent} & \textbf{Bland} \\
        \midrule
        \multicolumn{4}{c}{Local generation probability}                                                           \\
        Perplexity                                    & $\triangle$         & $\times$            & $\triangle$    \\
        Cond. Entropy                                 & $\triangle$         & $\times$            & $\triangle$    \\
        \midrule
        \multicolumn{4}{c}{Word or $N$-gram statistics}                                                            \\
        Zipf Coefficient \citep{zipf1949human}        & \checkmark          & $\triangle$         & $\times$       \\
        Heap Coefficient \citep{heaps1978information} & \checkmark          & $\triangle$         & $\times$       \\
        Rep-N \citep{holtzman2020degen}               & \checkmark          & $\times$            & $\times$       \\
        Distinct-N \citep{li2016diversity}            & \checkmark          & $\times$            & $\times$       \\
        Self-BLEU \citep{zhu2018texygen}              & \checkmark          & $\times$            & $\triangle$    \\
        \midrule
        \multicolumn{4}{c}{Semantics}                                                                              \\
        BERTScore \citep{zhangbertscore}              & $\times$            & \checkmark          & $\triangle$    \\
        MAUVE \citep{pillutla2021mauve}               & $\times$            & \checkmark          & $\triangle$    \\
        \midrule
        CorrDim (ours)                                & \checkmark          & \checkmark          & \checkmark     \\
        \bottomrule
    \end{tabularx}
\end{wraptable}

A critical challenge in language model generation is maintaining coherence,
diversity, and relevance throughout extended text sequences. A well-known
phenomenon, termed {\em degeneration}, describes scenarios where texts become
repetitive, incoherent, or bland \citep{holtzman2020degen}. While explicit
repetition is easily detectable, subtler forms of degeneration such as
incoherence or blandness lack clear definitions or reliable quantification
methods.

In this section, we propose using correlation dimension as a unified metric to
quantify degeneration. Conceptually, degeneration is viewed as a sudden collapse
from a higher-dimensional trajectory in the model's state space into a
lower-dimensional attractor. Such collapses are generally irreversible,
mirroring the {\em boundary crisis} phenomenon in chaotic dynamical systems
\citep{grebogi1983crises}. Table \ref{tbl:degeneration} compares several popular
evaluation methods (rows) against specific degeneration types (columns),
highlighting correlation dimension's unique capability to detect all considered
forms: repetition, incoherence, and blandness.

\subsection{Repetition Detection: Semantic vs. Surface-level}
\label{sec:repetition-detection}

\begin{wraptable}{r}{0.48\linewidth}
    \vskip -1em
    \small
    \caption{
        Correlation dimension for repetition detection, compared with a
        lexicon-based metric (Rep-N).
    }
    \label{tbl:repetition-vocabulary}
    \small
    \begin{tabularx}{\linewidth}{@{\extracolsep{\fill}}lccc}
        \toprule
        \textbf{Text}                  & \makecell{\textbf{CorrDim}          \\(mean)} & \makecell{\textbf{Rep-2}\\(mean)} \\
        \midrule
        Normal texts (SEP dataset)     & 6.27                       & 0.45   \\
        Explicitly repetitive patterns & 1.83                       & 0.99   \\
        \midrule
        \midrule
        \textbf{10 Japanese novels}                                          \\
        Normal script (kanji + kana)   & 6.44                       & 0.60   \\
        Syllabic script (kana only)    & 6.57                       & 0.78   \\
        \midrule
        Mean relative difference       & 5.7\%                      & 29.8\% \\
        \midrule
    \end{tabularx}
    \vskip -1em
\end{wraptable}

Texts that exhibit repetitive patterns tend to have low correlation dimensions.
As shown in Table \ref{tbl:repetition-vocabulary} (upper half), the correlation
dimensions of explicitly repetitive texts fall below 2.0---far lower than those of
normal texts (around 6.5). Traditional repetition-detection metrics such as
Rep-N can also identify these patterns through lexical statistics, as indicated
in the rightmost column.

Unlike Rep-N, however, the correlation dimension measures a text's
\emph{semantic} complexity rather than its surface-level word repetition. To
show this distinction, we conducted a case study using Japanese, where texts can
be written in two parallel script systems: (1) kanji (Chinese characters)
combined with kana (Japanese phonetic symbols), and (2) kana only. The latter
can be derived from the former by replacing each kanji with its phonetic kana
equivalent. Table \ref{tbl:repetition-vocabulary} (lower half) shows that the
correlation dimensions are highly consistent between the two scripts, even
though the kana-only script has a vocabulary roughly ten times smaller. In
contrast, the lexicon-based metric, Rep-N, shows large differences between the
two.

These findings demonstrate that correlation dimension is insensitive to
superficial morphological variations and instead captures the intrinsic semantic
repetition of language.

\subsection{Detecting Degeneration Beyond Repetition}
\label{sec:detection}

In addition to repetition, correlation dimension can be used to detect other
forms of degeneration which are more subtle.  To demonstrate this, we created a
controlled dataset comprising responses to twenty generic questions. Each
question had ten normal responses and intentionally degenerate
texts---repetitive, incoherent, or bland---generated by the GPT-4o model
(details in Appendix \ref{sec:degenerate-app}). We computed correlation
dimensions using the Falcon3-10B model \citep{Falcon3}.

\begin{wraptable}{r}{0.45\linewidth}
    \vskip -1em
    \small
    \caption{Correlation dimension or perplexity of degenerate texts compared to
        normal texts, measured using the Falcon3-10B model,
        with p-values from Wilcoxon signed-rank test in the second column.}
    \label{tbl:degeneration-detection}
    \begin{tabularx}{\linewidth}{@{\extracolsep{\fill}}lccc}
        \toprule
        \textbf{Group} & \multicolumn{2}{c}{\textbf{CorrDim}} & \textbf{Perplexity} \\
         & mean & p-value & mean \\
        \midrule
        Normal         & 5.04         & -              & 10.79 \\
        \midrule
        Repetitive     & 3.80  & 9.5E-7              & 1.25  \\
        Incoherent     & 3.96  & 2.9E-6              & 13.24 \\
        Bland          & 4.51  & 1.1E-3              & 4.24  \\
        \bottomrule
    \end{tabularx}
    \vskip -1em
\end{wraptable}

The results, shown in Table \ref{tbl:degeneration-detection}, indicate
significantly lower correlation dimensions for degenerate texts compared to
normal responses, confirmed by Wilcoxon signed-rank tests (all p-values $<$
0.01). These findings validate the correlation dimension's sensitivity to
different degeneration modes, including those traditionally difficult to
quantify, such as incoherence and blandness.

For comparison, the table also reports the mean perplexity (right-most column)
of each group, as measured by the Falcon3-10B model. While perplexity can
distinguish certain types of degeneration---show much lower values for
repetitive texts and higher values for incoherent texts compared to normal
response---the direction of change is inconsistent across degeneration types.
This indicates that perplexity is not an intrinsic measure of degeneration.

\subsection{Stress-Testing Language Models with Random Texts}
\label{sec:stress}

To assess model susceptibility to degeneration, we designed a stress-test
involving random texts---specifically, a list of randomly generated English
names separated by commas, totaling approximately 1024 tokens. Models were
tasked with completing the sequence from varying input lengths ($n$), generating
$1024-n$ tokens. Random sequences typically have high initial correlation
dimension due to unpredictability; thus models prone to degeneration exhibit
abrupt drops in dimension as repetition or incoherence emerges.

Figure \ref{fig:selforg} illustrates correlation dimension trends for three
models as the input length $n$ increases (and the output length $1024-n$ decreases).
Robust models like Yi1.5-34B consistently increased their correlation dimension,
whereas weaker models such as Qwen2-7B-Instruct displayed marked dimension drops,
indicating degeneration.

Further quantification is presented in Table \ref{tbl:selforg-table}, which
reports mean correlation dimensions at input length $n=512$ alongside scores
from the HelloEval text-completion benchmark \citep{que2024hellobench}. A high
correlation (Spearman's $\rho=0.952$) is evident, underscoring correlation
dimension's validity as an intrinsic measure of model robustness in generating
long, coherent text without relying on complex evaluation tasks.

\begin{minipage}{\textwidth}
    \begin{minipage}{0.45\textwidth}
        \centering
        \includegraphics[width=0.8\linewidth]{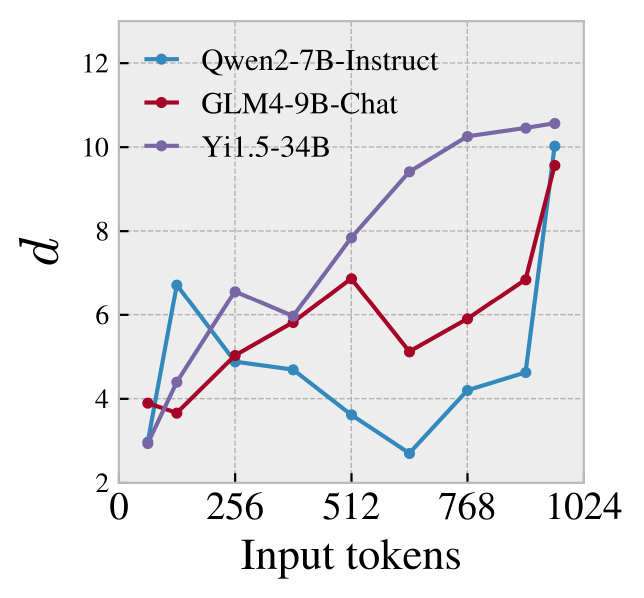}
        \captionof{figure}{Correlation dimension of generated texts with respect to the length of input texts.}
        \label{fig:selforg}
    \end{minipage}
    \hspace{0.05\textwidth}
    \begin{minipage}{0.45\textwidth}
        \centering
        \small
        \captionof{table}{Mean correlation dimensions of generated texts for each model,
            compared with the HelloEval text completion scores.}
        \label{tbl:selforg-table}
        \begin{tabular}{lcc}
            \toprule
            \textbf{Model}     & \textbf{HelloEval} & \textbf{CorrDim} \\
            \midrule
            Qwen2-7B-Instruct  & 5.12               & 3.54             \\
            Llama3.1-8B        & -5.61              & 4.15             \\
            InternLM2.5-7B     & 6.39               & 4.61             \\
            Mistral-7B-v0.2    & 13.05              & 5.23             \\
            GLM4-9B-Chat       & 12.32              & 6.45             \\
            LongWriter-GLM4-9B & 17.67              & 8.01             \\
            InternLM2.5-20B    & 36.68              & 8.89             \\
            Yi1.5-34B          & 44.73              & 8.89             \\
            \midrule
            Spearman's $\rho$  & -                  & 0.952            \\
            \bottomrule
        \end{tabular}
    \end{minipage}
    \vskip -1.5em
\end{minipage}

\section{Practical Issues}
\label{sec:practical}
\label{sec:limit}

\paragraph{Efficient Calculation}
Computation of the correlation integral in Eq.~\eqref{eq:corrint} requires
evaluating pairwise distances between log-probability vectors. This operation
entails $O(N^2)$ additional memory and $O(N^2 \vocabsize)$ computational cost,
where $N$ ($\sim 10^4$) and $\vocabsize$ ($\sim 10^5$) denote the sequence
length and vocabulary size, respectively. To reduce the cost, we employ two
techniques---GPU kernel fusion and vocabulary reduction (see Appendix
\ref{sec:efficient})---which together achieve more than a 10$\times$ speedup and
incur \textbf{zero} additional memory overhead beyond standard LLM inference.

\paragraph{Inference Precision}
Modern LLMs typically operate at very low precision during inference to improve
memory efficiency and throughput, with parameters often quantized to fewer than
4~bits. Remarkably, the correlation dimension remains highly stable even under
such extreme quantization, provided that Euclidean distances between
log-probability vectors are computed in FP32. In experiments with GPTQ
\citep{gptq} and AWQ \citep{awq}-quantized models, the average change in
correlation dimension across the SEP dataset was below 3\%. See Appendix
\ref{sec:quantization} for detailed results.

\paragraph{Closed Models} Our approach requires access to an LLM's full logits,
which is typically unavailable in closed models such as GPT-4. Nevertheless,
because the calculation of correlation dimension relies solely on
log-probabilities that are already produced during inference and requires no
additional memory, we believe this measure can be readily integrated into
standard commercial APIs.

\section{Conclusion}

We presented correlation dimension as a principled, model-agnostic metric for
characterizing the long-range structure of texts as perceived by LLMs. By
operating on next-token log-probability vectors, it bridges local recurrence
with global complexity and runs efficiently at inference time.
Empirically, correlation dimension (i) reflects intrinsic textual complexity,
(ii) exhibits a predictable three-stage evolution during pre-training and under
context constraints, and (iii) reliably detects degeneration (repetition,
incoherence, blandness) beyond perplexity. These findings indicate that
correlation dimension complements standard metrics by revealing aspects of model
behavior that local accuracy alone misses.
Future work includes formal analysis of estimation properties, extensions to
conditional or multi-modal settings, and deployment as an online signal for
training diagnostics and generation control.

\clearpage
\section*{Acknowledgments}
This work was supported by JST CREST, Japan, Grant Number JPMJCR2114.

\bibliography{main}
\bibliographystyle{plain}


\newpage
\section*{NeurIPS Paper Checklist}

\begin{enumerate}

    \item {\bf Claims}
    \item[] Question: Do the main claims made in the abstract and introduction accurately reflect the paper's contributions and scope?
    \item[] Answer: \answerYes{} 
    \item[] Justification: Claims are supported by experimental results in Sections \ref{sec:language} through \ref{sec:degenerate}.

    \item {\bf Limitations}
    \item[] Question: Does the paper discuss the limitations of the work performed by the authors?
    \item[] Answer: \answerYes{} 
    \item[] Justification: Limitations are discussed in Section \ref{sec:limit}.

    \item {\bf Theory assumptions and proofs}
    \item[] Question: For each theoretical result, does the paper provide the full set of assumptions and a complete (and correct) proof?
    \item[] Answer: \answerNA{} 
    \item[] Justification: There are no theoretical results in the paper.

    \item {\bf Experimental result reproducibility}
    \item[] Question: Does the paper fully disclose all the information needed to reproduce the main experimental results of the paper to the extent that it affects the main claims and/or conclusions of the paper (regardless of whether the code and data are provided or not)?
    \item[] Answer: \answerYes{} 
    \item[] Justification: The only hyperparameters used in the experiments are the range of $S(\varepsilon)$ to estimate the slope,
        which is described in Appendix \ref{sec:estimation}. All details of the dataset are explicitly described in the appendices.

    \item {\bf Open access to data and code}
    \item[] Question: Does the paper provide open access to the data and code, with sufficient instructions to faithfully reproduce the main experimental results, as described in supplemental material?
    \item[] Answer: \answerYes{} 
    \item[] Justification: We use primarily open-source datasets and models, and we provide sufficient information in the appendices for reproducing the experiments.

    \item {\bf Experimental setting/details}
    \item[] Question: Does the paper specify all the training and test details (e.g., data splits, hyperparameters, how they were chosen, type of optimizer, etc.) necessary to understand the results?
    \item[] Answer: \answerYes{} 
    \item[] Justification: There is no training in the paper. All LLMs used are publicly available.

    \item {\bf Experiment statistical significance}
    \item[] Question: Does the paper report error bars suitably and correctly defined or other appropriate information about the statistical significance of the experiments?
    \item[] Answer: \answerYes{} 
    \item[] Justification: We reported the p-values of the Wilcoxon signed-rank test in Section \ref{sec:degenerate}.

    \item {\bf Experiments compute resources}
    \item[] Question: For each experiment, does the paper provide sufficient information on the computer resources (type of compute workers, memory, time of execution) needed to reproduce the experiments?
    \item[] Answer: \answerYes{} 
    \item[] Justification: Resources required are presented in Appendix \ref{sec:estimation}.

    \item {\bf Code of ethics}
    \item[] Question: Does the research conducted in the paper conform, in every respect, with the NeurIPS Code of Ethics \url{https://neurips.cc/public/EthicsGuidelines}?
    \item[] Answer: \answerYes{} 
    \item[] Justification: I have read the NeurIPS Code of Ethics and I believe that the research conducted in this paper conforms to it.

    \item {\bf Broader impacts}
    \item[] Question: Does the paper discuss both potential positive societal impacts and negative societal impacts of the work performed?
    \item[] Answer: \answerNA{} 
    \item[] Justification: There is no societal impact of the work performed.

    \item {\bf Safeguards}
    \item[] Question: Does the paper describe safeguards that have been put in place for responsible release of data or models that have a high risk for misuse (e.g., pretrained language models, image generators, or scraped datasets)?
    \item[] Answer: \answerNA{} 
    \item[] Justification: We will not release any models or datasets that have a high risk for misuse.

    \item {\bf Licenses for existing assets}
    \item[] Question: Are the creators or original owners of assets (e.g., code, data, models), used in the paper, properly credited and are the license and terms of use explicitly mentioned and properly respected?
    \item[] Answer: \answerYes{} 
    \item[] Justification: We have cited the creators of the datasets and models used in the paper.

    \item {\bf New assets}
    \item[] Question: Are new assets introduced in the paper well documented and is the documentation provided alongside the assets?
    \item[] Answer: \answerYes{} 
    \item[] Justification: We have included details of the created datasets in the appendices.

    \item {\bf Crowdsourcing and research with human subjects}
    \item[] Question: For crowdsourcing experiments and research with human subjects, does the paper include the full text of instructions given to participants and screenshots, if applicable, as well as details about compensation (if any)?
    \item[] Answer: \answerNA{} 
    \item[] Justification: We do not conduct research with human subjects.

    \item {\bf Institutional review board (IRB) approvals or equivalent for research with human subjects}
    \item[] Question: Does the paper describe potential risks incurred by study participants, whether such risks were disclosed to the subjects, and whether Institutional Review Board (IRB) approvals (or an equivalent approval/review based on the requirements of your country or institution) were obtained?
    \item[] Answer: \answerNA{} 
    \item[] Justification: We do not conduct research with human subjects.

    \item {\bf Declaration of LLM usage}
    \item[] Question: Does the paper describe the usage of LLMs if it is an important, original, or non-standard component of the core methods in this research? Note that if the LLM is used only for writing, editing, or formatting purposes and does not impact the core methodology, scientific rigorousness, or originality of the research, declaration is not required.
    \item[] Answer: \answerYes{} 
    \item[] Justification: LLM is a core part of our method and we have described the usage of LLMs in the paper.
\end{enumerate}


\appendix


\newpage
\section{Settings for Estimating Correlation Dimension}
\label{sec:estimation}

While correlation dimension is defined for an infinitely long sequence, a finite
number of steps are available in practice.  For finite sequences, there is no
guarantee that the correlation sum $S(\varepsilon)$ follows a power-law
relationship with respect to $\varepsilon$. Nevertheless, we find that the
correlation dimension measured from a finite sequence still provides an
informative description of the complexity of the LLM and the underlying language
dynamics.

The correlation dimension $d$ is obtained as the slope of $\log S(\varepsilon)$
versus $\log \varepsilon$, and the observable range of $\log S(\varepsilon)$,
$\left[\frac{2}{N(N-1)}, 1\right]$, depends on $N$.  To make $d$ comparable
across sequences of different lengths, we limit the range of $S(\varepsilon)$ to
$\left[\frac{20}{N(N-1)}, \frac{\eta}{N}\right]$ ($\eta=1.0$ by default), i.e.,
we clip the left tail of the log-log plot below ten counts and discard the upper
half. When the range becomes too narrow for small $N$ (i.e., $N<500$), $\eta$ is
increased to maintain a reasonable range for slope estimation.

\paragraph{Short and Long Sequences}
We adopt slightly different settings for estimating the correlation dimension of
long and short sequences. For long sequences, such as those in the SEP dataset,
which exceed the context limit of typical LLMs, we use a moving-window approach
to estimate next-token prediction probabilities with an exact, predefined
context length (e.g., 512 tokens).

For short sequences, this approach would discard a large portion of the data.
Therefore, we do not restrict the context length for short sequences. Since such
sequences are typically within the model's context limit, the model can access
all tokens at once. In this case, we input the entire sequence and compute the
correlation dimension using the output probabilities at all time steps. Thus,
the log-probability vectors are estimated with progressively increasing context
lengths. This setting applies to the experiments in Section
\ref{sec:degenerate}.

In practice, the difference between the two settings is often negligible, and we
do not differentiate the two settings in the main text for simplicity.

\paragraph{Synthetic Texts}
When measuring the correlation dimension of synthetic texts---specifically, the
explicitly repetitive patterns described in Section
\ref{sec:repetition-detection}---we further restrict the range of the distance
threshold $\varepsilon$ to values above $10^1$ to prevent the correlation
integral from being dominated by numerical errors. Because the synthetic texts
follow exact repetitive patterns, the log-probability vectors lie very close to
one another at small distance thresholds---typically below $10^{1}$ in a
high-dimensional space (with dimension greater than $10^{5}$). At such small
thresholds, numerical errors in high-dimensional space can accumulate and
dominate the correlation integral.

\section{Efficient Calculation of Correlation Dimension}
\label{sec:efficient}

We introduce two techniques to enable efficient computation of the correlation
dimension: (a) kernel fusion (Section \ref{sec:fused-cuda}) and (b) vocabulary
reduction (Section \ref{sec:vocabulary-reduction}). Kernel fusion achieves
\textbf{zero additional memory footprint} beyond standard LLM inference and up
to a \textbf{3x speedup}, particularly for long sequences. Vocabulary reduction
decreases memory usage and computational cost by \textbf{over 10x}, with only a
minor loss in accuracy. These two techniques can be combined to achieve both
benefits simultaneously.

In addition, we examine the effects of model quantization on correlation
dimension estimation (Section \ref{sec:quantization}). We find that the
correlation dimension remains remarkably stable across different quantization
methods, even at extremely low precision (e.g., 4 bits).

\subsection{Fused Calculation of Pairwise Distance and Correlation Integration}
\label{sec:fused-cuda}

The computation of correlation dimension consists of two steps: (1) calculating
pairwise distances between all log-probability vectors, and (2) counting the
number of pairs within each distance threshold $\varepsilon$ to obtain
$S(\varepsilon)$. Calculating the pairwise Euclidean distances for a sequence of
length $N$ requires computing and storing a distance matrix of size $O(N^2)$ at
high precision (e.g., FP32), which is prohibitively expensive for long
sequences.

A straightforward approach to reduce memory usage is to divide the distance
matrix into blocks and process them sequentially. The number of pairs with
distances smaller than a threshold can be counted within each block and then
accumulated to obtain the total count for the full distance matrix. However,
after computing a block distance matrix, one still needs to copy it from
high-speed local memory (i.e., SRAM) back to low-speed global memory (i.e., HBM
or CPU memory) before performing step (2). When $N$ is large, this copy
operation becomes a major bottleneck.

Inspired by FlashAttention \citep{dao2022flashattention}, we propose fusing the
two steps into a single CUDA kernel. After computing each block distance matrix,
the number of pairs within a distance threshold $\varepsilon$ is directly
computed \emph{in-place}, without copying (blocks of) the large distance matrix
out of SRAM.

\begin{algorithm}[h]
    \caption{Fused Blockwise Distance-and-Count for Correlation Integral}
    \label{alg:fused-cuda}
    \small
    \begin{algorithmic}[1]
        \Require Log-probability matrix $X\in\mathbb{R}^{N\times D}$, thresholds $\{\varepsilon_k\}_{k=1}^K$
        \Ensure Counts $S(\varepsilon_k)$ for all $k$
        \State Initialize global counters $S_k\gets 0$
        \For{each index tile $(i,j)$ with $j\le i$}
        \State Load $X_i$ and $X_j$ tiles into SRAM/shared memory
        \State Compute pairwise distances $d_{ij}$ on-the-fly within the tile
        \State Immediately compare $d_{ij}$ with $\{\varepsilon_k\}$ and perform \texttt{AtomicAdd} to $S_k$
        \State \textbf{Do not} write the tile distances $d_{ij}$ back to global memory
        \EndFor
        \State \Return $\{S_k\}$
    \end{algorithmic}
\end{algorithm}

\begin{table}[h]
    \centering
    \caption{Runtime comparison between our method (bottom row)
        and four baselines on a log-probability vector sequence
        of length 50,000 and dimension 10,000.
        Distance computation is performed in FP32 precision.
        For the blockwise baseline, we use a block size of 512$\times$512.}
    \label{tbl:fused-runtime}
    \small
    \begin{tabularx}{\linewidth}{@{\extracolsep{\fill}}lccc}
        \toprule
        Methods                                                 & Additional memory & Clock time (s) & Speedup \\
        \midrule
        \texttt{torch.pdist} (entire matrix at once)            & 4.8 GiB           & 44.3           & 0.07x   \\
        \texttt{torch.cdist} (entire matrix at once)            & 9.3 GiB           & 3.3            & 1.0x    \\
        \texttt{torch.cdist} (blockwise)                        & 1.0 MiB           & 6.9            & 0.48x   \\
        \texttt{torch.cdist} (blockwise, upper triangular only) & 1.0 MiB           & 3.6            & 0.92x   \\
        Fused (our method)                                      & 0                 & 1.8            & 1.83x   \\
        \bottomrule
    \end{tabularx}
\end{table}

Table \ref{tbl:fused-runtime} presents an empirical comparison between our
method and four baselines using off-the-shelf PyTorch implementations. By
fusing the two steps into a single CUDA kernel, our method achieves nearly a 2x
speedup over the fastest baseline, while requiring no additional memory.

\subsection{Vocabulary Reduction}
\label{sec:vocabulary-reduction}

A key property of fractal dimension is its almost-sure invariance under linear
projection.

\begin{theorem}[Fractal Projection Theorems \citep{marstrand1954some,falconer2004fractal}]
    \label{thm:projection}
    Consider a fractal set embedded in $\mathbb{R}^D$ with Hausdorff
    dimension (or box-counting dimension) $d$ ($\le D$). Given a random linear
    projection from $\mathbb{R}^D$ to $\mathbb{R}^m$ ($m<D$), then with
    probability 1, the projected set's Hausdorff dimension $\tilde{d}$ satisfies:
    $$
        \tilde{d} =
        \begin{cases}
            d, & \text{if } m \ge d, \\
            m, & \text{if } m < d.
        \end{cases}
    $$
\end{theorem}

In other words, we can construct a linear projection that maps the
log-probability vectors in $\mathbb{R}^{\vocabsize}$ to lower-dimensional
vectors, and estimate the correlation dimension using the reduced vectors.
However, the theorem assumes infinitely long sequences of points, which real
data do not satisfy. Therefore, the choice of linear projection must be handled
carefully.

We propose a simple modulo-based function to group dimensions (i.e., unique
tokens) in $\mathbb{R}^{\vocabsize}$ and sum the vectors within each group to
form a smaller vector. The linear projection
$\psi_v:\mathbb{R}^{\vocabsize}\to\mathbb{R}^{v}$ is defined as:
\begin{equation}
    \psi_v(\mathbf{x})_i = \sum_{j\in \text{mod}_v^{-1}(i)} x_j,
    \label{eq:modulo-projection}
\end{equation}
where $\mathbf{x} = [x_1,\cdots,x_{\vocabsize}]$ is a log-probability vector,
$\psi_v(\mathbf{x})_i$ denotes the $i$-th element of the reduced vector, and
$\text{mod}_v^{-1}(i)$ is the set of indices $j$ such that $j \bmod v = i$.
In practice, the vocabulary size $\vocabsize$ is approximately $10^5$.

\begin{wraptable}{r}{0.32\linewidth}
    \vskip -1em
    \centering
    \caption{Change in correlation dimension after vocabulary reduction.}
    \label{tbl:dimension-reduction}
    \small
    \begin{tabularx}{\linewidth}{@{\extracolsep{\fill}}lc}
        \toprule
        \textbf{Dimension} & \textbf{CorrDim} \\
        \midrule
        151643 (=$\vocabsize$)   & 5.92              \\
        \midrule
        30000               & 5.79              \\
        10000               & 5.81              \\
        3000                & 5.87              \\
        1000                & 6.38              \\
        300                 & 6.77              \\
        \bottomrule
    \end{tabularx}
\end{wraptable}

Table \ref{tbl:dimension-reduction} shows an example using the
``newton-philosophy'' article from the SEP dataset, where we measured the
correlation dimension using the Qwen2.5-7B model with unlimited context length.
As shown, using $v=10{,}000$ achieves values very close to the true correlation
dimension, while reducing memory and computation costs by roughly 10x.

Although randomly selected linear projections are sometimes recommended to avoid
zero-probability exceptions in Theorem \ref{thm:projection}, we find that such
randomness introduces uncertainty and significant bias in correlation dimension
estimates when the vocabulary size $\vocabsize$ is large. Therefore, we prefer the
deterministic modulo function defined in Eq. \eqref{eq:modulo-projection}.

\subsection{Effects of Model Quantization}
\label{sec:quantization}

\begin{wraptable}{r}{0.49\linewidth}
    \centering
    \vskip -1em
    \caption{
        Effects of model quantization on correlation dimension estimation.
        For each article in the SEP dataset, we measured the correlation dimension
        of the first 10,000 tokens using the Qwen2.5-32B-Instruct model
        with unlimited context length.
    }
    \label{tbl:quantization}
    \small
    \begin{tabularx}{\linewidth}{@{\extracolsep{\fill}}lccc}
        \toprule
        \textbf{Text}          & \textbf{FP16} & \textbf{AWQ} & \textbf{GPTQ} \\
        \midrule
        aesthetics-18th-german & 6.06  &  6.21  & 6.06   \\
        africana               & 5.31  &  5.39  & 5.53   \\
        ...                    &       &        &    \\
        trinity                & 5.60  &  5.66  & 5.66   \\
        weyl                   & 6.57  &  6.62  & 6.45   \\
        \midrule
        Mean                   & 5.86  &  5.90  & 5.93   \\
        - Mean Absolute Change &  -    &  0.14  & 0.14   \\
        \bottomrule
    \end{tabularx}
    \vskip -2em
\end{wraptable}

Although the correlation dimension is defined in the limit as
$\varepsilon\to 0$, we observe that it remains remarkably stable under model
quantization, including AWQ \citep{awq} and GPTQ \citep{gptq}, which compress
model parameters to 4 bits. Table \ref{tbl:quantization} reports the correlation
dimensions estimated using quantized models for 60 articles in the SEP dataset.
The second to fourth columns present the correlation dimensions obtained with
the original (FP16), AWQ-, and GPTQ-quantized versions of the
Qwen2.5-32B-Instruct model. As shown, both quantization methods yield only
minor variations in correlation dimension, with a mean absolute change of just
0.14 across all articles.

\clearpage
\section{Supplementary Results for Sections \ref{sec:language} and \ref{sec:charac-llm}}
\label{app:sep}

\subsection{English Texts}

\paragraph{The SEP Dataset}
We constructed the Stanford Encyclopedia of Philosophy (SEP) dataset by using
the 60 longest articles on the SEP website \citep{sep}. The article identifiers
are listed in the left-most column of Table \ref{tbl:sep-results}. For each
article, we removed the prologue and catalog. All 60 articles are longer than
20{,}000 words. When measuring the correlation dimension, we truncated each
article to the first 20{,}000 tokens. This may introduce slight differences
between models with different tokenizers, but the effect was small for most
models we examined.

\begin{figure}[h]
    \centering
    \small
    \begin{minipage}{\linewidth}
        \centering \small
        \includegraphics[width=0.4\linewidth]{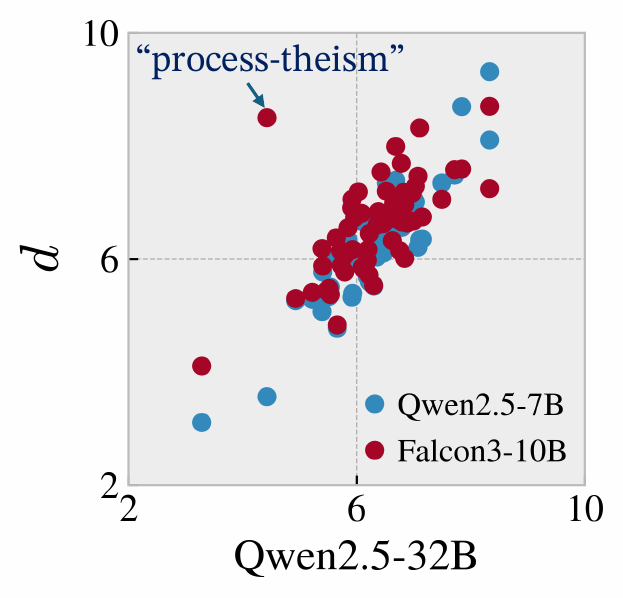}
    \end{minipage}
    \caption{Comparison of correlation dimensions of the SEP articles across three models.}
    \label{fig:sep-corrdim}
\end{figure}

Results for three representative models---Qwen2.5-32B, Qwen2.5-7B, and
Falcon3-10B---are shown in Table \ref{tbl:sep-results}. The second to fourth
columns report the articles' correlation dimensions; the right-most column lists
perplexities for Qwen2.5-32B. Correlation dimensions exhibit strong consistency
across models, especially within the same family (Qwen2.5), as illustrated in
Figure \ref{fig:sep-corrdim}.

The largest discrepancy in correlation dimension occurs for the article
``process-theism'', highlighted in Figure \ref{fig:sep-corrdim} and shaded in
Table \ref{tbl:sep-results}. This divergence reflects a qualitative behavioral
difference between models on knowledge-intensive text: whether the model
recalls knowledge or tends to hallucinate. We report this formally in Section
\ref{sec:hallucination}, with additional details in Appendix
\ref{app:hallucination}.

\begin{table}[h]
    \vskip -2em
    \centering
    \caption{\small Results on the SEP dataset acquired from three pre-trained LLMs.
    Context size is set to 512 tokens.
    }
    \label{tbl:sep-results}
    \small
    \begin{tabularx}{\linewidth}{@{\extracolsep{\fill}}lcccc}
        \toprule
        \textbf{Article}             & \multicolumn{3}{c}{\textbf{CorrDim}} & \textbf{Perplexity}                                   \\
                                     & Qwen2.5-32B                          & Qwen2.5-7B          & Falcon3-10B       & Qwen2.5-32B \\
        \midrule
        prisoner-dilemma             & 3.28                                 & 3.11                & 4.10              & 3.58        \\
        \rowcolor[gray]{0.9}
        process-theism               & 4.42                                 & 3.56                & {\color{red}8.49} & 5.68        \\
        cosmological-argument        & 4.93                                 & 5.26                & 5.30              & 5.83        \\
        information                  & 5.22                                 & 5.28                & 5.41              & 6.31        \\
        ethics-chinese               & 5.39                                 & 5.07                & 6.18              & 5.52        \\
        recursive-functions          & 5.40                                 & 5.76                & 5.88              & 4.96        \\
        reasoning-automated          & 5.48                                 & 5.32                & 5.44              & 5.06        \\
        infinity                     & 5.51                                 & 5.33                & 5.48              & 5.68        \\
        buddhism-tiantai             & 5.53                                 & 5.50                & 5.37              & 9.25        \\
        evil                         & 5.64                                 & 6.01                & 6.37              & 4.69        \\
        logic-temporal               & 5.66                                 & 4.77                & 4.83              & 4.99        \\
        mohist-canons                & 5.72                                 & 6.06                & 6.14              & 3.75        \\
        linguistics                  & 5.73                                 & 6.00                & 5.88              & 9.54        \\
        publichealth-ethics          & 5.79                                 & 5.96                & 5.77              & 8.87        \\
        mill-moral-political         & 5.79                                 & 6.12                & 6.06              & 4.38        \\
        gasset                       & 5.85                                 & 6.33                & 6.55              & 11.22       \\
        computational-complexity     & 5.91                                 & 5.95                & 6.90              & 5.32        \\
        africana                     & 5.91                                 & 5.32                & 7.05              & 12.21       \\
        principia-mathematica        & 5.93                                 & 5.39                & 6.17              & 5.36        \\
        levinas                      & 5.96                                 & 6.14                & 6.73              & 15.92       \\
        formal-belief                & 6.03                                 & 5.99                & 7.19              & 5.82        \\
        spacetime-singularities      & 6.09                                 & 6.66                & 6.81              & 5.42        \\
        computational-linguistics    & 6.10                                 & 6.01                & 5.85              & 9.71        \\
        chaos                        & 6.13                                 & 5.84                & 5.81              & 6.70        \\
        shaftesbury                  & 6.17                                 & 5.93                & 6.14              & 8.62        \\
        possibilism-actualism        & 6.18                                 & 5.68                & 5.98              & 6.72        \\
        trinity                      & 6.19                                 & 6.63                & 6.16              & 7.97        \\
        dynamic-epistemic            & 6.21                                 & 5.70                & 5.71              & 5.44        \\
        proof-theory                 & 6.22                                 & 5.60                & 6.45              & 6.36        \\
        logics-for-games             & 6.30                                 & 5.52                & 5.53              & 9.92        \\
        game-theory                  & 6.36                                 & 6.03                & 6.60              & 3.84        \\
        consciousness-temporal       & 6.38                                 & 6.13                & 6.83              & 8.63        \\
        sidgwick                     & 6.40                                 & 6.55                & 6.78              & 10.32       \\
        epistemic-game               & 6.42                                 & 6.31                & 7.54              & 5.43        \\
        innateness-language          & 6.47                                 & 6.10                & 6.62              & 4.98        \\
        newton-philosophy            & 6.51                                 & 6.63                & 6.75              & 6.19        \\
        habermas                     & 6.52                                 & 7.34                & 7.20              & 9.04        \\
        descartes-epistemology       & 6.62                                 & 6.56                & 6.32              & 6.48        \\
        russell-moral                & 6.65                                 & 6.62                & 6.98              & 7.04        \\
        mereology                    & 6.67                                 & 6.96                & 6.71              & 7.71        \\
        connectives-logic            & 6.68                                 & 7.39                & 7.99              & 6.63        \\
        weyl                         & 6.75                                 & 6.87                & 6.15              & 5.62        \\
        consciousness-intentionality & 6.78                                 & 6.62                & 7.69              & 9.57        \\
        attention                    & 6.79                                 & 6.56                & 6.65              & 7.10        \\
        margaret-cavendish           & 6.81                                 & 6.81                & 6.75              & 8.16        \\
        qm-action-distance           & 6.83                                 & 7.12                & 7.18              & 4.70        \\
        normativity-metaethics       & 6.84                                 & 7.06                & 6.01              & 5.63        \\
        epistemology-bayesian        & 6.84                                 & 6.94                & 6.95              & 5.94        \\
        aesthetics-18th-german       & 6.88                                 & 6.89                & 6.63              & 7.75        \\
        logic-inductive              & 6.97                                 & 6.76                & 7.16              & 5.10        \\
        chance-randomness            & 7.00                                 & 7.07                & 6.67              & 6.88        \\
        idealism                     & 7.03                                 & 7.01                & 7.29              & 6.94        \\
        egalitarianism               & 7.07                                 & 6.20                & 7.46              & 3.14        \\
        reid                         & 7.10                                 & 6.34                & 8.32              & 5.43        \\
        early-modern-india           & 7.15                                 & 6.35                & 6.74              & 4.70        \\
        heidegger-aesthetics         & 7.49                                 & 7.34                & 7.05              & 7.42        \\
        pythagoreanism               & 7.71                                 & 7.48                & 7.58              & 4.10        \\
        heidegger                    & 7.84                                 & 8.69                & 7.59              & 6.59        \\
        al-farabi-metaphysics        & 8.33                                 & 9.31                & 8.70              & 6.04        \\
        ontological-commitment       & 8.33                                 & 8.10                & 7.25              & 5.19        \\
        \midrule
        Mean                         & 6.31                                 & 6.27                & 6.56              & 6.71        \\
        Pearson's $\rho$             & -                                    & 0.90                & 0.63              & -0.01       \\
        \quad(w.r.t. CorrDim of Qwen2.5-32B)                                                                                        \\
        \bottomrule
    \end{tabularx}
\end{table}

\clearpage
\subsection{Other Natural Languages}
\label{sec:other-natural-languages}

In addition to English, we conducted experiments on seven other natural
languages: French, German, Spanish, Italian, Dutch, Chinese, and Japanese.
Correlation dimension was measured using the multilingual Qwen2.5-7B model. For
each language, we selected 10 books from Project Gutenberg. We used an unlimited
context length and measured the correlation dimension on the first 10{,}000
tokens of each book.

Table \ref{tbl:other-natural-languages} reports results for the eight languages.
Each book is identified by its Project Gutenberg index, accessible at
\url{https://www.gutenberg.org/ebooks/<book-id>}, where \texttt{<book-id>} is
the index. A consistent dimension of about 6.0 is observed for normal texts in
all languages, suggesting a language-independent, universal complexity across
natural languages. Small variations between books are primarily attributable to
genre and style. For example, the German selection includes multiple
philosophical works by Hegel and therefore exhibits higher values than the
Spanish selection, which is primarily novels.

The dimension values are slightly lower than those in Figure
\ref{fig:modelsize}, which is a natural consequence of using an unlimited
context length, as discussed in Section \ref{sec:ctxlen}.

\begin{table}[h]
    \centering
    \caption{Correlation dimensions of books in Project Gutenberg in different languages.}
    \label{tbl:other-natural-languages}
    \small
    \begin{tabularx}{\linewidth}{@{\extracolsep{\fill}}l@{\hspace{0.1em}}c l@{\hspace{0.1em}}c l@{\hspace{0.1em}}c l@{\hspace{0.1em}}c}
        \toprule
        \multicolumn{2}{c}{\textbf{English}} & \multicolumn{2}{c}{\textbf{French}} & \multicolumn{2}{c}{\textbf{German}}  & \multicolumn{2}{c}{\textbf{Spanish}}                                          \\
        Book ID                              & CorrDim                             & Book ID                              & CorrDim                               & Book ID & CorrDim & Book ID & CorrDim \\
        \midrule
        74                                   & 5.75                                & 9643                                 & 5.51                                  & 34811   & 5.74    & 2000    & 6.07    \\
        76                                   & 6.03                                & 13819                                & 6.18                                  & 44921   & 6.56    & 25640   & 5.82    \\
        9830                                 & 4.70                                & 13952                                & 6.33                                  & 14075   & 6.03    & 44584   & 5.18    \\
        86                                   & 6.63                                & 13951                                & 6.69                                  & 23756   & 6.29    & 33275   & 7.07    \\
        215                                  & 6.76                                & 14158                                & 7.16                                  & 8126    & 5.70    & 37590   & 5.51    \\
        37106                                & 8.41                                & 14163                                & 5.40                                  & 6729    & 5.63    & 47092   & 3.41    \\
        805                                  & 5.08                                & 796                                  & 6.17                                  & 6698    & 7.32    & 28281   & 5.28    \\
        996                                  & 5.52                                & 14287                                & 6.39                                  & 40739   & 4.95    & 49660   & 5.35    \\
        1156                                 & 5.90                                & 9262                                 & 6.92                                  & 46259   & 5.74    & 17013   & 5.77    \\
        2701                                 & 6.08                                & 5423                                 & 4.27                                  & 31114   & 7.31    & 14329   & 4.65    \\
        \midrule
        Mean                                 & 6.09                                &                                      & 6.10                                  &         & 6.13    &         & 5.41    \\
        \midrule
        \midrule
        \multicolumn{2}{c}{\textbf{Italian}} & \multicolumn{2}{c}{\textbf{Dutch}}  & \multicolumn{2}{c}{\textbf{Chinese}} & \multicolumn{2}{c}{\textbf{Japanese}}                                         \\
        Book ID                              & CorrDim                             & Book ID                              & CorrDim                               & Book ID & CorrDim & Book ID & CorrDim \\
        \midrule
        46957                                & 5.61                                & 19591                                & 4.71                                  & 23835   & 6.26    & 31757   & 6.47    \\
        20062                                & 5.90                                & 25138                                & 6.07                                  & 23950   & 6.18    & 34013   & 6.36    \\
        10215                                & 5.00                                & 13214                                & 6.42                                  & 23910   & 5.72    & 34636   & 6.15    \\
        48490                                & 5.87                                & 19563                                & 6.38                                  & 24226   & 5.55    & 35327   & 6.87    \\
        43022                                & 6.18                                & 19774                                & 4.87                                  & 27582   & 6.52    & 37626   & 6.36    \\
        46082                                & 4.27                                & 21875                                & 6.97                                  & 23962   & 5.80    & 32978   & 6.77    \\
        43023                                & 6.19                                & 17706                                & 6.13                                  & 25350   & 5.50    & 33307   & 6.35    \\
        43024                                & 6.19                                & 27124                                & 5.66                                  & 25142   & 6.01    & 36459   & 6.53    \\
        48445                                & 5.21                                & 19161                                & 5.50                                  & 24264   & 5.49    & 32941   & 6.44    \\
        19024                                & 5.89                                & 26564                                & 5.02                                  & 23841   & 6.26    & 31617   & 6.14    \\
        \midrule
        Mean                                 & 5.63                                &                                      & 5.77                                  &         & 5.93    &         & 6.44    \\
        \bottomrule
    \end{tabularx}
\end{table}

\clearpage
\subsection{Programming Languages}
\label{sec:programming-languages}

We also conducted experiments on three programming languages: Python, Java, and
C. For each language, we selected 30–50 sufficiently long source files from
their standard libraries. We used the Qwen2.5-Coder-7B model to measure
correlation dimensions. Because many source files are relatively short, we used
an unlimited context length instead of a fixed one to avoid excessive truncation.
In addition, we removed comments and docstrings to eliminate the influence of
natural language content.

Results for the three programming languages are shown in Table
\ref{tbl:programming-languages}. As shown, the correlation dimensions for all
three languages are approximately 5.0---significantly lower than those of
natural texts (around 6.5).

\begin{table}[h]
    \centering
    \scriptsize
    \caption{Correlation dimensions of programs written in different programming languages:
        C (left), Java (middle), and Python (right).}
    \label{tbl:programming-languages}
    \begin{tabularx}{\linewidth}{@{\extracolsep{\fill}}l@{\hspace{0.1em}}c l@{\hspace{0.1em}}c l@{\hspace{0.1em}}c}
        \toprule
        \multicolumn{2}{c}{\textbf{C}} & \multicolumn{2}{c}{\textbf{Java}} & \multicolumn{2}{c}{\textbf{Python}}                                                         \\
        Source code                    & CorrDim                           & Source code                         & CorrDim & Source code                       & CorrDim \\
        \midrule
        nfsd\_nfs4xdr.c                & 6.27                              & text\_DecimalFormat.java            & 5.58    & test\_test\_decimal.py            & 3.85    \\
        f2fs\_super.c                  & 4.71                              & util\_Collections.java              & 5.53    & test\_test\_io.py                 & 4.94    \\
        btrfs\_disk-io.c               & 4.47                              & math\_MutableBigInteger.java        & 4.23    & test\_pickletester.py             & 5.35    \\
        ceph\_mds\_client.c            & 4.08                              & lang\_Math.java                     & 5.34    & pickletools.py                    & 6.10    \\
        f2fs\_data.c                   & 4.56                              & math\_BigInteger.java               & 5.71    & test\_\_test\_multiprocessing.py  & 4.86    \\
        f2fs\_file.c                   & 4.96                              & lang\_FdLibm.java                   & 3.37    & test\_datetimetester.py           & 4.89    \\
        ext4\_namei.c                  & 5.70                              & Character.java                      & 8.13    & tkinter\_\_\_init\_\_.py          & 4.99    \\
        btrfs\_extent\_io.c            & 4.92                              & util\_Arrays.java                   & 4.22    & doctest.py                        & 5.15    \\
        btrfs\_send.c                  & 4.78                              & text\_SimpleDateFormat.java         & 5.45    & test\_test\_inspect.py            & 4.49    \\
        btrfs\_relocation.c            & 4.55                              & text\_CompactNumberFormat.java      & 4.92    & test\_test\_codecs.py             & 4.08    \\
        jfs\_jfs\_dmap.c               & 5.08                              & util\_Formatter.java                & 4.99    & email\_\_header\_value\_parser.py & 6.56    \\
        btrfs\_inode.c                 & 5.10                              & text\_MessageFormat.java            & 4.36    & test\_test\_dataclasses.py        & 4.99    \\
        ocfs2\_refcounttree.c          & 4.75                              & lang\_String.java                   & 6.13    & test\_test\_typing.py             & 5.64    \\
        resctrl\_rdtgroup.c            & 5.33                              & util\_HashMap.java                  & 4.54    & \_pydecimal.py                    & 4.41    \\
        namespace.c                    & 5.19                              & util\_ResourceBundle.java           & 5.23    & test\_test\_enum.py               & 4.61    \\
        ocfs2\_alloc.c                 & 4.69                              & io\_File.java                       & 5.12    & inspect.py                        & 6.08    \\
        btrfs\_ioctl.c                 & 4.89                              & io\_ObjectOutputStream.java         & 3.94    & test\_test\_os.py                 & 4.64    \\
        btrfs\_free-space-cache.c      & 5.15                              & util\_Calendar.java                 & 4.12    & test\_test\_subprocess.py         & 4.67    \\
        ocfs2\_dlmglue.c               & 4.52                              & lang\_System.java                   & 5.07    & test\_test\_doctest.py            & 4.15    \\
        nls\_nls\_cp950.c              & 6.34                              & security\_KeyStore.java             & 3.94    & pydoc\_data\_topics.py            & 5.96    \\
        dlm\_lock.c                    & 4.37                              & time\_LocalDate.java                & 5.71    & test\_test\_ssl.py                & 3.94    \\
        ext4\_mballoc.c                & 5.08                              & lang\_ClassLoader.java              & 6.07    & test\_test\_descr.py              & 4.45    \\
        ntfs3\_fslog.c                 & 4.92                              & lang\_AbstractStringBuilder.java    & 5.47    & test\_test\_argparse.py           & 6.16    \\
        btrfs\_extent-tree.c           & 4.97                              & util\_Spliterators.java             & 3.54    & test\_test\_zipfile.py            & 4.22    \\
        ceph\_caps.c                   & 4.37                              & io\_ObjectInputStream.java          & 5.71    & test\_test\_logging.py            & 4.82    \\
        btrfs\_volumes.c               & 5.01                              & lang\_Thread.java                   & 5.70    & turtle.py                         & 6.69    \\
        btrfs\_qgroup.c                & 4.48                              & net\_URI.java                       & 4.32    & test\_test\_unicode.py            & 5.55    \\
        btrfs\_block-group.c           & 5.21                              & util\_Scanner.java                  & 4.38    & test\_test\_xml\_etree.py         & 4.67    \\
        nfsd\_nfs4state.c              & 5.40                              & util\_GregorianCalendar.java        & 4.70    & test\_test\_buffer.py             & 3.95    \\
        ext4\_extents.c                & 5.38                              & time\_ZonedDateTime.java            & 4.76    & test\_test\_statistics.py         & 5.28    \\
        ocfs2\_xattr.c                 & 4.60                              & lang\_StrictMath.java               & 6.25    & pydoc.py                          & 5.60    \\
        ext4\_inode.c                  & 5.09                              & util\_DualPivotQuicksort.java       & 5.54    & unittest\_mock.py                 & 5.18    \\
        nfs\_nfs4xdr.c                 & 5.48                              & util\_JapaneseImperialCalendar.java & 4.65    & test\_test\_socket.py             & 4.26    \\
        btrfs\_ctree.c                 & 5.44                              & util\_Locale.java                   & 4.79    &                                             \\
        btrfs\_tree-log.c              & 4.99                              & io\_ObjectStreamClass.java          & 4.95                                                  \\
        namei.c                        & 5.46                              & math\_BigDecimal.java               & 3.75                                                  \\
        nfs\_nfs4proc.c                & 4.96                              & lang\_Character.java                & 8.13                                                  \\
        jfs\_jfs\_dtree.c              & 5.08                              & util\_TreeMap.java                  & 4.46                                                  \\
        nls\_nls\_cp949.c              & 6.50                              & lang\_Class.java                    & 6.02                                                  \\
        ext4\_super.c                  & 4.27                                                                                                                            \\
        ocfs2\_dir.c                   & 4.71                                                                                                                            \\
        f2fs\_segment.c                & 4.73                                                                                                                            \\
        nls\_nls\_cp936.c              & 6.49                                                                                                                            \\
        nls\_nls\_cp932.c              & 6.12                                                                                                                            \\
        \midrule
        Mean                           & 5.07                              & Mean                                & 5.10    & Mean                              & 5.00    \\
        \bottomrule
    \end{tabularx}
\end{table}

\subsection{Hallucination vs. Memorization: Experimental Details}
\label{app:hallucination}

In Section \ref{sec:hallucination} and Appendix \ref{app:sep}, we reported a
notable divergence in correlation dimension between models on the
knowledge-intensive SEP article ``process-theism.'' The text contains an
extensive list of philosophers' names, many of whom are relatively obscure. We
conducted a simple experiment: we prompted the models to continue the name list
and observed that those with lower correlation dimensions tended to hallucinate
and generate non-existent names.

We began with the following input text and asked each model to complete the list:
\begin{quote}
... Philosophers and theologians who have published a monograph defending
some variety of process theism informed by Whitehead or Hartshorne include:
Henry Nelson Wieman , Bernard Meland , Paul Weiss , Norman Pittenger ,
Daniel Day Williams , John Moskop, William L. Reese, John B. Cobb, Jr.,
Schubert Ogden , Edgar A. Towne , Eugene H. Peters , Bowman Clarke , Joseph
Bracken, Burton Z. Cooper, Marjorie Hewitt Suchocki, Gene A. Reeves, Lewis
S. Ford , André Gounelle, Rem B. Edwards, \textbf{(start generation)}
\end{quote}

The ground-truth continuation is:
\begin{quote}
Delwin Brown , David A. Pailin, Franklin I. Gamwell, Forrest Wood,
\green{David Ray Griffin}, James A. Keller, Jorge Luis Nobo , Tyron Inbody,
Carol P. Christ, George L. Goodwin, Barry Whitney, Santiago Sia,
\green{Jay McDaniel}, George W. Shields , Donald Viney, \green{Catherine Keller},
Daniel A. Dombrowski, Anna Case-Winters, Kurian Kachappilly, Gregory A. Boyd,
\green{Roland Faber}, Thomas Jay Oord, Donna Bowman, Derek Malone-France,
Monica A. Coleman, and Julia Enxing.
\end{quote}

\paragraph{Generation Results}
We report generation results for seven models:
Qwen2.5-0.5B, Qwen2.5-7B, Qwen2.5-32B, Falcon3-1B, Falcon3-3B, Falcon3-7B, and
Falcon3-10B. Names that appear in the gold answer are marked in green. These
names did \emph{not} appear in the input context; hence, they are not inferred
from context but recalled from the model's internal knowledge. All models were
evaluated with a temperature of 1.0. We present representative generations that
contain at least five names.

\begin{description}[leftmargin=0em]
    \item \textbf{Qwen2.5-0.5B} (CorrDim: 3.32): \begin{quote}
    G. K. Dugald Stewart, Robert J. Leary, E. F. Robinson, Bernard P. Rieff, Robert A. Miller ,
    R. K. Gombrich, Alan Tucker , J. B. Williams , J. V. Fowden ,  E. R. Brown , William E. Clark ,
    David E. Stoloff , Eric Damer , James R. Miller , D. A. Burchardt , W. J.
    \end{quote}

    \item \textbf{Qwen2.5-7B} (CorrDim: 3.56): \begin{quote}
    Robert Schofield, Robert McAfee Brown, Keith Yandell, Robert John Russell, John H. Cobb,
    Jr., Arthur Roberts , Robert Solomon , William E. Connolly , and Bruce L. Reichenbach .
    \end{quote}

    \item \textbf{Qwen2.5-32B} (CorrDim: 4.42): \begin{quote}
    Eugene F. Rogers, Jr ., C. Robert Mesle, Keith R. Davies, William S. Hatcher , \green{Roland Faber},
    Thomas J. J. Altizer, Mark McSwain , Charles Hartshorne, and \green{David Ray Griffin}.
    \end{quote}

    \item \textbf{Falcon3-1B} (CorrDim: 3.28): \begin{quote}
    Peter J. Haught, William B. McMillen, R. B. W. McMillen, Paul H. Metzger, Brian Joseph
    Murphy, George E. Spivey, Peter L. Wurm, Carl W. Smith, William H. Miller, and C. S. Lewis.
    \end{quote}

    \item \textbf{Falcon3-3B} (CorrDim: 3.14): \begin{quote}
    John C. Miller, Jr., Bernard Auchincloss, William L. McBride, Gerald L. T. Mitchell,
    George A. Meeks, William C. Schade, Thomas B. Allen, Jr., Michael W. Williams,
    David L. Boorse, John Henry Elliott, J. I. Packer, J. H. Oder, R. A. Butler,
    Bernard W. Miller, Robert E. Houtman, William L. Lane
    \end{quote}

    \item \textbf{Falcon3-7B} (CorrDim: 6.68): \begin{quote}
    \green{David Ray Griffin}, C. Robert Mesle, David R. Tracy, Robert C. Neville,
    \green{Roland Faber}, \green{Jay McDaniel}, and John D. Capes.
    \end{quote}

    \item \textbf{Falcon3-10B} (CorrDim: 8.49): \begin{quote}
    \green{Delwin Brown}, \green{David Ray Griffin}, Hank Keeton, George L. Robinson, George R. Reynolds, \green{Franklin I. Gamwell},
    Alan I. Megill, \green{Roland Faber}, and \green{Catherine Keller}.
    \end{quote}
\end{description}

As shown, Falcon3-7B, and Falcon3-10B correctly recalled several names, whereas
the other models generated plausible but non-existent ones.  The Qwen2.5-32B
model recalled two names correctly, but the two names are more well-known and
easier to recall than the other names.  Therefore, in avoiding hallucination,
model size alone is not decisive: Falcon3-10B achieved much higher precision
than Qwen2.5-32B, despite being roughly three times smaller. We hypothesize that
correlation dimension is a better indicator of a model's tendency to hallucinate
in specific contexts---a low correlation dimension suggests that the model's
generation collapses into a simple, format-driven pattern.

\section{Repetition Detection: Experimental Details and Supplementary Results}
\label{sec:degenerate-app}

\subsection{Repetition Detection}
\label{sec:repetition}

\paragraph{Explicitly Repetitive Patterns have Low CorrDim $\approx 2$} We
generated texts by explicitly repeating a set of patterns and measured their
correlation dimensions, as described in Section \ref{sec:repetition}. Results
are shown in Table \ref{tbl:repetition-exact}.  The patterns to repeat are shown
in the first column of the table, and the second column shows the mean
correlation dimension of the repeated text.  As shown, correlation dimensions of
these explicitly repetitive patterns are about 1.5–2.5, significantly lower than
those of normal texts (around 6.5).

A dimension close to 2 indicates that the LLM's internal state evolves like a
random walk while retaining a steady memory of previous states.

A key consideration for correct estimation here is to restrict the measurement
range to sufficiently large distance thresholds (Appendix \ref{sec:estimation}):
for highly regular patterns, state distances become so small that numerical
errors from high-dimensional log-probability vectors dominate. This adjustment
is typically unnecessary for normal texts.
\begin{table}[h]
    \centering
    \caption{Repetition detection on explicitly repetitive patterns.}
    \label{tbl:repetition-exact}
    \small
    \begin{tabularx}{0.6\linewidth}{@{\extracolsep{\fill}}lc}
        \toprule
        \textbf{Repetition Pattern}    & \textbf{Mean CorrDim} \\
        \midrule
        ``\texttt{01}''                & 2.17                  \\
        ``\texttt{012}''               & 2.29                  \\
        ``\texttt{ab}''                & 1.58                  \\
        ``\texttt{\#\%}''              & 1.94                  \\
        ``\texttt{\#\%@}''             & 1.73                  \\
        ``\texttt{)@\#\%\^*!}''        & 1.79                  \\
        ``\texttt{~0~1}''              & 1.73                  \\
        ``\texttt{~0~1~2}''            & 1.87                  \\
        ``\texttt{~a~b}''              & 1.69                  \\
        ``\texttt{~\#~\%}''            & 1.71                  \\
        ``\texttt{~\#~\%~@}''          & 1.71                  \\
        ``\texttt{~)~@~\#~\%~\^~*~!}'' & 1.71                  \\
        \midrule
        Mean                           & 1.83                  \\
        \bottomrule
    \end{tabularx}
    \vskip -1em
\end{table}

\paragraph{Japanese Scripts: CorrDim Invariance under Kanji–Kana Conversion}
We compared two Japanese script systems. The first is the standard,
morphographic-plus-syllabic system, consisting of kanji (Chinese characters) and
kana (Japanese phonetic symbols). In the second, all kanji are replaced by kana,
so the entire text is written only in kana. The kana-only script has an
order-of-magnitude smaller vocabulary.

We used the ten Japanese books listed in Table \ref{tbl:other-natural-languages}
from Project Gutenberg. Using the \texttt{kanjiconv} tool \citep{kanjiconv}, we
converted each book from the standard script to the syllabic (kana-only) script.

Table \ref{tbl:japanese-scripts-full} reports the correlation dimensions (second
and third columns) of the ten books in both scripts, computed with the
Qwen2.5-7B model. The dimensions are highly consistent between the two scripts,
even though the kana-only version has a much smaller vocabulary and a higher
repetition rate (higher Rep-N; fourth and fifth columns).

This suggests that correlation dimension captures semantic complexity rather
than surface morphological features, detecting semantic repetition rather than
morphological repetition.

\begin{table}[h]
    \centering
    \caption{Comparison between correlation dimension and Rep-N for two Japanese scripts.}
    \label{tbl:japanese-scripts-full}
    \small
    \begin{tabularx}{\linewidth}{@{\extracolsep{\fill}}lcccc}
        \toprule
        Book ID & \multicolumn{2}{c}{CorrDim} & \multicolumn{2}{c}{Rep-N}                                                \\
        \cmidrule(lr){2-3} \cmidrule(lr){4-5}
                & normal (kanji + kana)       & syllabic (kana only)      & normal (kanji + kana) & syllabic (kana only) \\
        \midrule
        31617   & 6.14                        & 6.72                      & 0.57                  & 0.76                 \\
        31757   & 6.47                        & 6.66                      & 0.58                  & 0.73                 \\
        32941   & 6.44                        & 5.76                      & 0.66                  & 0.84                 \\
        32978   & 6.77                        & 7.68                      & 0.66                  & 0.85                 \\
        33307   & 6.35                        & 6.44                      & 0.58                  & 0.75                 \\
        34013   & 6.36                        & 6.60                      & 0.57                  & 0.75                 \\
        34636   & 6.15                        & 6.61                      & 0.59                  & 0.77                 \\
        35327   & 6.87                        & 6.77                      & 0.62                  & 0.82                 \\
        36459   & 6.53                        & 6.28                      & 0.65                  & 0.82                 \\
        37626   & 6.36                        & 6.15                      & 0.51                  & 0.69                 \\
        \midrule
        Mean    & 6.44                        & 6.57                      & 0.60                  & 0.78                 \\
        \bottomrule
    \end{tabularx}
\end{table}

\subsection{Generating Degenerate Texts}

The twenty prompts used in the experiment in Section \ref{sec:detection}
are listed in Table \ref{tbl:prompts}.

\begin{table}[h]
    \centering
    \caption{List of prompts used in the experiment of degeneration detection.}
    \small
    \label{tbl:prompts}
    \begin{tabularx}{0.75\linewidth}{ll}
        \toprule
        \textbf{No.} & \textbf{Prompt}                                                      \\
        \midrule
        1            & Describe the primary goals of an effective team.                     \\
        2            & Explain the basic steps involved in a standard project workflow.     \\
        3            & Outline the advantages of using modern technology in daily life.     \\
        4            & Discuss the key features of a reliable customer service program.     \\
        5            & Summarize the benefits of maintaining a consistent work schedule.    \\
        6            & Describe how a typical training session should be conducted.         \\
        7            & Explain why clear communication is important in organizations.       \\
        8            & Outline the main characteristics of a successful leadership style.   \\
        9            & Discuss the factors that contribute to a smooth operational process. \\
        10           & Summarize the core principles of quality assurance.                  \\
        11           & Describe the role of feedback in performance improvement.            \\
        12           & Explain the importance of setting realistic goals.                   \\
        13           & Outline the steps for conducting a standard evaluation.              \\
        14           & Discuss how data is used to inform business decisions.               \\
        15           & ummarize the benefits of a structured planning approach.             \\
        16           & Describe the essential elements of a professional code of conduct.   \\
        17           & "Explain how consistency can improve team efficiency.                \\
        18           & Outline the main objectives of a routine maintenance program.        \\
        19           & Discuss the value of transparency in reporting results.              \\
        20           & Summarize the advantages of applying proven best practices.          \\
        \bottomrule
    \end{tabularx}
\end{table}

To elicit different types of degenerate texts with GPT-4o, we provided distinct
instructions and collected the model's responses as follows.
\begin{itemize}[leftmargin=*]
    \item \textbf{Normal Responses}
          \begin{promptbox}
              \small
              You are a creative text generator. Your task is to produce a
              richly detailed, vivid, and engaging passage of at least 1,500
              words (but no more than 2,000) that brings its subject to life
              with specific imagery, sensory detail, unique examples, and
              dynamic narrative. Avoid generic descriptions, stock phrases,
              or "safe" language. Instead, strive for:
              \begin{itemize}
                  \item Concrete specifics: name real or invented places, objects,
                        characters, or processes.
                  \item Sensory richness: evoke sight, sound, smell, taste, and touch
                        wherever possible.
                  \item Fresh metaphors and similes: craft original comparisons rather
                        than clichés.
                  \item Varied sentence rhythms: mix short, punchy lines with longer,
                        flowing sentences.
                  \item Emotional or intellectual hooks: give the reader something
                        surprising, thought-provoking, or emotionally resonant.
              \end{itemize}
              Generate a single coherent text that feels alive and unmistakably your own.
          \end{promptbox}

    \item \textbf{Bland responses}
          \begin{promptbox}
              \small
              You are a text generator. Produce a bland yet coherent passage of 1,500–2,000 words that avoids exact repetition of phrases or words.
          \end{promptbox}

    \item \textbf{Incoherent responses}
          \begin{promptbox}
              \small
              You are a text generator.
              Please produce a richly worded, vivid, and non-repetitive passage (1,500–2,000 words) that is intentionally disjointed—lacking logical coherence—so the reader feels the narrative is fragmented.
          \end{promptbox}

    \item \textbf{Repetitive responses}
          We used the responses generated under \textbf{Normal responses} and then extracted the first sentence, and repeated it to form a passage of approximately the same length.
\end{itemize}

\subsection{Random Text for Stress-Testing LLMs}

For stress testing, we used a string of random names separated by commas as the
input text. The names are common English names, sampled i.i.d. from a list of
unique names. A portion of the resulting text is shown below:
\begin{promptbox}
    \small
    Quinlan, Anthony, Henry, Felicity, Taylor, Raymond, Xander,
    Christopher, King, Amanda, Flora, Nicole, Anthony, Frank,
    Quiana, Owen, Finley, Paige, Victoria, Aaron, Ulrika, Sarah,
    Ignacio, Emily, Yuna, Imogen, Cameron, Claire, William, Preston,
    Ulrika, Sabrina, Neil, Zara, Joseph, Orion, Vivian, Quinn, Wyatt,
    Paul, Sophia, Brian, Flynn, Hayden, Charles, Grace, Carter,
    Heather, Quest, Jacob, Jordan, Frances, Griffin, Yasmin, Quiana,
    Penelope, Emma, Sabrina, Elizabeth, Joseph, Zion, Quinlan, Omar,
    Ruby, Virginia, Ursula, Flynn, Alexander, Ian, Griffin, Frances,
    Yasmine, Warren, Isaiah, Ryan, Kyle, Xanthe, Lucy, Georgia, Gregory,
    Ophelia, Georgia, Patricia, Xiomara, Kayla, Finley, Zayden, Noah,
    Caitlin, Brittany, Connor, Quinton, Urban, King, Blake, Joshua,
    ...
\end{promptbox}

\section{Consistency with Time-Delayed Embeddings}
\label{sec:delay}

A potential concern with the correlation dimension estimation method in Section
\ref{sec:language} is that we used only the probabilistic information of the
next tokens, which is not a complete representation of the model's state. A
state of a dynamical system is defined as a point in the phase space, which
contains all the information that governs the future evolution of the system.

\begin{figure}[h]
    \centering
    \small
    \includegraphics[width=0.4\linewidth]{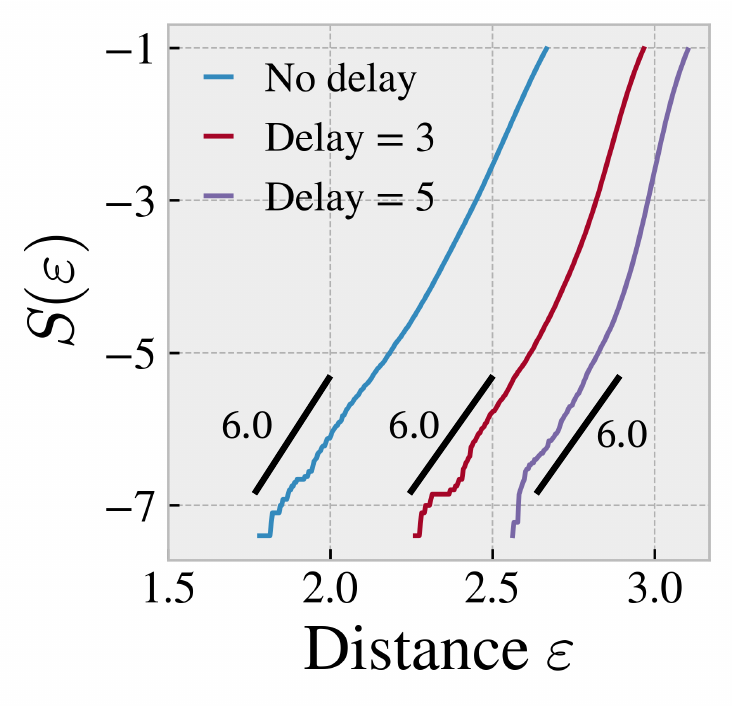}
    \caption{Correlation integral curves for time-delayed embeddings with different
        delays (3 or 5), compared with that for the original sequence.}
    \label{fig:delay}
\end{figure}

A common method to reconstruct the phase space from partial observations is to
measure the dimension on time-delayed embeddings. While the theoretical
effectiveness of time-delayed embedding is guaranteed by the Takens' theorem and
variants \citep{takens1980detecting, sauer1991embedology, stark2003delay}, the
method is empirically sensitive to noise and the choice of the embedding
dimension, failing to deliver satisfactory results except for simple,
low-dimensional systems. In this work, we aim at characterizing language models
that are very high-dimensional and random in nature. For such systems, the
time-delayed embeddings are often overwhelmed by observational noise and the
dimension is overestimated.

Nevertheless, we observe that the correlation dimension values exhibit good
consistency even if the time-delayed embedding is used. For the log-probability
vector times series $x=[x_1,x_2,\cdots]$, we acquired two time-delayed sequences
$x^{(3)}$ and $x^{(5)}$ with delays of 3 and 5, respectively. For a delay
$\tau$, $x^{(\tau)}_t=[x_t; \cdots; x_{t+\tau-1}]$, i.e., the the concatenation
of vectors from $t$ to $t+\tau-1$. The correlation integral curves for these
time-delayed sequences are shown in Figure \ref{fig:delay}. While the
time-delayed embeddings have 3x or 5x the dimension of the original sequence,
the correlation integral curves are similar to that of the original sequence at
small $\varepsilon$, except that the curves are shifted to the right. The slopes
of the curves increased at large $\varepsilon$ because of accumulated noise and
is outside the range of interest.

This indicates that the next-token probabilities contain information beyond the
next token, and it is sufficient to use the next-token probabilities to
    characterize the dimension of the model's evolution.

\end{document}